\PassOptionsToPackage{unicode}{hyperref}
\PassOptionsToPackage{hyphens}{url}
\ifx\pdfoutput\undefined\else\pdfoutput=1\fi
\PassOptionsToPackage{dvipsnames,svgnames,x11names}{xcolor}
\documentclass[
  12pt]{article}

\usepackage{amsmath,amssymb}
\usepackage{amsthm}
\usepackage{iftex}
\usepackage{soul} 
\usepackage{xcolor}
\usepackage{changes} 
\usepackage{algorithmicx} 
\usepackage{algpseudocode} 
 
\usepackage{float}    
\usepackage{todonotes}
\usepackage{placeins} 
\usepackage{multirow}
\ifPDFTeX
  \usepackage[T1]{fontenc}
  \usepackage[utf8]{inputenc}
  \usepackage{textcomp} 
\else 
  \usepackage{unicode-math}
  \defaultfontfeatures{Scale=MatchLowercase}
  \defaultfontfeatures[\rmfamily]{Ligatures=TeX,Scale=1}
\fi
\usepackage{lmodern}
\ifPDFTeX\else  
\fi
\IfFileExists{upquote.sty}{\usepackage{upquote}}{}
\IfFileExists{microtype.sty}{
  \usepackage[]{microtype}
  \UseMicrotypeSet[protrusion]{basicmath} 
}{}
\makeatletter
\@ifundefined{KOMAClassName}{
  \IfFileExists{parskip.sty}{%
    \usepackage{parskip}
  }{
    \setlength{\parindent}{0pt}
    \setlength{\parskip}{6pt plus 2pt minus 1pt}}
}{
  \KOMAoptions{parskip=half}}
\makeatother
\usepackage{xcolor}
\makeatletter
\ifx\textbf\undefined\else
  \let\oldparagraph\textbf
  \renewcommand{\textbf}{
    \@ifstar
      \xxxParagraphStar
      \xxxParagraphNoStar
  }
  \newcommand{\xxxParagraphStar}[1]{\oldparagraph*{#1}\mbox{}}
  \newcommand{\xxxParagraphNoStar}[1]{\oldparagraph{#1}\mbox{}}
\fi
\ifx\subparagraph\undefined\else
  \let\oldsubparagraph\subparagraph
  \renewcommand{\subparagraph}{
    \@ifstar
      \xxxSubParagraphStar
      \xxxSubParagraphNoStar
  }
  \newcommand{\xxxSubParagraphStar}[1]{\oldsubparagraph*{#1}\mbox{}}
  \newcommand{\xxxSubParagraphNoStar}[1]{\oldsubparagraph{#1}\mbox{}}
\fi
\makeatother

\usepackage{longtable,booktabs,array}
\usepackage{calc} 
\usepackage{etoolbox}
\makeatletter
\patchcmd\longtable{\par}{\if@noskipsec\mbox{}\fi\par}{}{}
\makeatother
\IfFileExists{footnotehyper.sty}{\usepackage{footnotehyper}}{\usepackage{footnote}}
\makesavenoteenv{longtable}
\usepackage{graphicx}
\makeatletter
\def\maxwidth{\ifdim\Gin@nat@width>\linewidth\linewidth\else\Gin@nat@width\fi}
\def\maxheight{\ifdim\Gin@nat@height>\textheight\textheight\else\Gin@nat@height\fi}
\makeatother
\setkeys{Gin}{width=\maxwidth,height=\maxheight,keepaspectratio}
\makeatletter
\def\fps@figure{htbp}
\makeatother

\makeatletter
\@ifpackageloaded{caption}{}{\usepackage{caption}}
\AtBeginDocument{%
\ifdefined\contentsname
  \renewcommand*\contentsname{Table of contents}
\else
  \newcommand\contentsname{Table of contents}
\fi
\ifdefined\listfigurename
  \renewcommand*\listfigurename{List of Figures}
\else
  \newcommand\listfigurename{List of Figures}
\fi
\ifdefined\listtablename
  \renewcommand*\listtablename{List of Tables}
\else
  \newcommand\listtablename{List of Tables}
\fi
\ifdefined\figurename
  \renewcommand*\figurename{Figure}
\else
  \newcommand\figurename{Figure}
\fi
\ifdefined\tablename
  \renewcommand*\tablename{Table}
\else
  \newcommand\tablename{Table}
\fi
}
\@ifpackageloaded{float}{}{\usepackage{float}}
\floatstyle{ruled}
\@ifundefined{c@chapter}{\newfloat{codelisting}{h}{lop}}{\newfloat{codelisting}{h}{lop}[chapter]}
\floatname{codelisting}{Listing}

\makeatother
\makeatletter
\@ifpackageloaded{caption}{}{\usepackage{caption}}
\@ifpackageloaded{subcaption}{}{\usepackage{subcaption}}
\makeatother

\usepackage{algorithm}
\usepackage{algpseudocode}
\usepackage{amsfonts}
\usepackage{amsthm}
\usepackage{placeins}
\newtheorem{theorem}{Theorem}

\newtheorem{assumption}{Assumption}
\newtheorem{proposition}{Proposition}
\usepackage{threeparttable}

\ifLuaTeX
  \usepackage{selnolig}  
\fi
\usepackage[]{natbib}
\usepackage{bookmark}

\IfFileExists{xurl.sty}{\usepackage{xurl}}{} 
\hypersetup{
  pdftitle={Tensor-Efficient High-Dimensional Q-learning},
  pdfauthor={Author 1; Author 2},
  pdfkeywords={reinforcement learning, tensor decomposition, Q-learning, sample efficiency, high-dimensional},
  colorlinks=true,
  linkcolor={blue},
  filecolor={Maroon},
  citecolor={Blue},
  urlcolor={Blue},
  pdfcreator={LaTeX via pandoc}}

\newcommand{\anon}{0}

\usepackage{float}
\begin{document}

\def\spacingset#1{\renewcommand{\baselinestretch}%
{#1}\small\normalsize} \spacingset{1}


\if0\anon
{
  \title{\bf Tensor-Efficient High-Dimensional $Q$-learning}
  \author{Junyi Wu\\
  Department of Industrial \& System Engineering\\
  University of Washington\\
  junyiwu@uw.edu
  \and
  Dan Li\\
  Department of Industrial \& System Engineering\\
  University of Washington\\
  dli27@uw.edu}
  \date{}
  \maketitle
} \fi

\bigskip
\begin{abstract}
High-dimensional reinforcement learning(RL) faces challenges with complex calculations and low sample efficiency in large state-action spaces. Q-learning algorithms struggle particularly with the curse of dimensionality, where the number of state-action pairs grows exponentially with problem size. While neural network-based approaches like Deep Q-Networks have shown success, they do not explicitly exploit problem structure. Many high-dimensional control tasks exhibit low-rank structure in their value functions, and tensor-based methods using low-rank decomposition offer parameter-efficient representations. However, existing tensor-based Q-learning methods focus on representation fidelity without leveraging this structure for exploration. We propose Tensor-Efficient Q-Learning (TEQL), which represents the Q-function as a low-rank CP tensor over discretized state-action spaces and exploits the tensor structure for uncertainty-aware exploration. TEQL incorporates Error-Uncertainty Guided Exploration (EUGE), which combines tensor approximation error with visit counts to guide action selection, along with frequency-aware regularization to stabilize updates. Under matched parameter budgets, experiments on classic control tasks demonstrate that TEQL outperforms both matrix-based low-rank methods and deep RL baselines in sample efficiency, making it suitable for resource-constrained applications where sampling costs are high.
\end{abstract}

\noindent%
\vfill

\newpage
\spacingset{1.8} 

\section{Introduction}\label{sec-intro}

Value function estimation is the basic computational challenge in reinforcement learning, where agents must evaluate the expected cumulative rewards associated with states or state-action pairs to guide optimal decision-making \citep{sutton2018reinforcement, bellman1957dynamic, bertsekas1996neuro}. The value function provides the fundamental link between observed rewards and long-term planning, with theoretical foundations rooted in dynamic programming and Markov Decision Processes \citep{puterman1994markov, bertsekas1996neuro, szepesvari2010algorithms}. Recent work has further revealed structural properties of value functions that can be exploited for efficient learning \citep{dadashi2019value, yang2020harnessing}.

This challenge becomes particularly acute in high-dimensional state-action spaces with inherent discrete structure, which arise naturally in operations research and industrial systems. In these domains, state and action spaces factorize into multiple discrete components, and the total number of configurations grows exponentially in the number of dimensions, creating severe computational and statistical burdens.
 
Two representative settings illustrate this structure. In clinical treatment optimization, the state of a patient is described by multiple discrete clinical indicators such as severity levels, organ function grades, and recovery stages, and the treatment decision is a combination of discrete choices among drugs, dosages, and interventions \citep{komorowski2018artificial, liu2019learning}. The state-action space grows combinatorially with the number of clinical dimensions, yet drug interactions are predominantly pairwise, with higher-order interactions among three or more treatments being rare in practice, which implies that the value function admits low-rank tensor structure. Each treatment trial involves a real patient, making data collection expensive and sample-efficient learning essential. In multi-echelon inventory management, the state is the vector of discrete inventory levels across all stocking locations, and the action is the joint replenishment decision specifying an integer order quantity at each location. The state-action space grows exponentially with the number of locations, yet the optimal policy often admits low-rank structure because demands are driven by a small number of shared factors such as seasonal trends and regional economic conditions \citep{powell2007approximate, gijsbrechts2022can}. Similar factored discrete structure arises in materials process optimization \citep{wu2011experiments}, multi-component maintenance scheduling \citep{de2020review}, network routing, and resource allocation.
 
When learning must proceed online through direct interaction with the environment, sample efficiency becomes critical. The curse of dimensionality renders classical tabular methods impractical \citep{bellman1957dynamic, szepesvari2010algorithms, powell2007approximate}, as each interaction may be costly or risky. These considerations highlight the need for approximation methods that are both expressive and statistically efficient \citep{gheshlaghi2013minimax, sam2023overcoming}.

Classical approaches range from tabular Q-learning \citep{watkins1992q} to linear function approximation \citep{bradtke1996linear} and deep neural networks such as DQN \citep{mnih2015human} and SAC \citep{christodoulou2019sacdiscrete}. While deep RL methods offer representational flexibility, they typically require extensive samples and do not explicitly exploit the multi-dimensional structure inherent in discretized state-action spaces, motivating structured alternatives.

Low-rank structure offers a principled alternative that addresses both sample efficiency and structural exploitation. A line of theoretical work has established that when the underlying MDP admits low-rank structure, sample complexity can be fundamentally reduced from scaling with the product of state and action space sizes to scaling with their sum \citep{jiang2017contextual, agarwal2020flambe, uehara2022representation}. This theoretical insight motivates methods that learn compact value function representations. Matrix-based approaches reduce parameters via SVD or nuclear-norm regularization \citep{shah2020sample}, and recent model-free methods such as LoRa-VI achieve finite-sample guarantees through structured estimation \citep{stojanovic2024lora, modi2024modelfree}. Tensor decomposition further extends this idea: by representing the Q-function as a multi-dimensional array with CP factorization, tensor methods reduce complexity from exponential to linear in the number of dimensions while preserving mode-wise interactions that matrix flattening discards \citep{tsai2021tensor, rozada2024tensor}.

However, existing tensor-based Q-learning focuses on representation fidelity rather than sample efficiency, employing standard $\varepsilon$-greedy exploration without exploiting structural uncertainty. Matrix-based methods with sample-efficiency guarantees, conversely, do not extend naturally to multi-dimensional state-action representations. To the best of our knowledge, no existing method combines low-rank tensor structure with uncertainty-aware exploration for sample-efficient online learning.

Based on this motivation, we propose Tensor-Efficient Q-Learning (TEQL), an online, model-free reinforcement learning framework that takes advantage of the low-rank tensor structure and uncertainty-aware exploration to improve the statistical efficiency of value-function estimation in high-dimensional settings. Our approach makes three main contributions.

First, we develop a low-rank tensor Q-learning scheme with frequency-based regularization that compresses the value-function representation from exponential to linear complexity in the number of dimensions. We establish convergence in expectation to a neighborhood of the optimal $Q$-function under a low-rank structural assumption, with the radius explicitly separating approximation and stochastic error components.

Second, we propose Error-Uncertainty Guided Exploration (EUGE), an uncertainty-aware action-selection rule that augments value estimates with a bonus based on decomposition error and visit counts. By tracking how rapidly the tensor approximation changes across iterations, EUGE prioritizes state-action pairs whose estimates remain uncertain and helps allocate samples more effectively in high-dimensional spaces.

Third, we provide empirical evaluations, including ablation and sensitivity studies, demonstrating that TEQL improves sample efficiency relative to tensor and non-tensor baselines under the same interaction budget. Across classical control environments, TEQL achieves faster improvement in returns and more rapid reduction of value-function error, and we examine how its performance varies with tensor rank, discretization levels, and exploration parameters.

\section{Preliminaries}\label{sec-prelim}

We review $Q$-learning, low-rank value function representations, and structural assumptions that support the TEQL framework. We also specify the notation used throughout this paper. Scalars use lowercase letters (e.g., $q, r, \gamma$), vectors use bold lowercase (e.g., $\mathbf{s}, \mathbf{a}, \mathbf{f}$), matrices use bold uppercase (e.g., $\mathbf{Q}, \boldsymbol{F}$), and tensors use calligraphic letters (e.g., $\mathcal{Q}, \mathcal{T}$). Sets use calligraphic letters (e.g., $\mathcal{S}, \mathcal{A}$); $|\mathcal{S}|$ 
and $|\mathcal{A}|$ denote the cardinalities of the discretized state and action 
spaces, respectively. The optimal Q-function is $q^*$, and $\hat{\mathcal{Q}}$ denotes the low-rank tensor approximation. Rank-$R$ refers to the CP rank of the low-rank tensor. Factor matrices are $\boldsymbol{F}_n$ with time-indexed versions $\boldsymbol{F}_n^{(t)}$. Transition and reward functions use $\mathcal{P}$ and $\mathcal{R}$.

The method developed in this paper primarily targets MDPs with discrete, factored state-action spaces, where each dimension of the state and action takes values from a finite set. This discrete structure is inherent in the application domains discussed in Section~\ref{sec-intro}, including clinical treatment optimization, inventory management, and maintenance scheduling. Many high-dimensional control tasks exhibit value functions with low-rank tensor structure, where state and action variables interact through a limited number of latent factors. Tensor decomposition provides a natural way to exploit this structure, and discrete indexing is the mechanism through which the factored state-action components are represented and compressed. Although continuous state-action spaces are not the primary target of this method, TEQL can still be applied to such settings by discretizing each continuous dimension into a finite set of bins prior to learning, which is a common practice in Q-learning. The approximation error introduced by this discretization step is absorbed into the constant $B_R$ analyzed in Section~\ref{subsec:structural_assumptions}.

\subsection{Low-Rank Value Function Representations for \texorpdfstring{$Q$}{Q}-Learning}\label{subsec:low-rank}

A discounted Markov Decision Process (MDP) is defined by the tuple $\langle \mathcal{S}, \mathcal{A}, \mathcal{P}, \mathcal{R}, \gamma \rangle$. For a state $s$ and action $a$, the optimal Q-function satisfies
\begin{equation}
q^*(s,a) = \mathbb{E}[r + \gamma \max_{a'} q^*(s',a') \mid s,a].
\end{equation}
$Q$-learning updates the estimate $q(s,a)$ as
\begin{equation}
q(s,a) \leftarrow q(s,a) + \alpha \bigl( r + \gamma \max_{a'} q(s',a') - q(s,a) \bigr).
\end{equation}
Tabular updates become infeasible in high-dimensional spaces, so compact representations such as low-rank matrix or tensor models are required.

The Q-function can be arranged as a matrix $\mathbf{Q} \in \mathbb{R}^{C_{\mathcal{S}} \times C_{\mathcal{A}}}$ or as a tensor when the state and action contain multiple components. For high-dimensional state-action spaces, a tensor representation preserves the multiway structure.

Let $s \in \mathbb{R}^{D_S}$ and $a \in \mathbb{R}^{D_A}$ denote underlying continuous 
state and action variables when applicable. After discretization, the combined index 
vector $(i_1,\ldots,i_N)$ with $N = D_S + D_A$ identifies one tensor entry. This 
discretized representation naturally induces a multi-dimensional tensor form of the 
Q-function, given by
\begin{equation}
\mathcal{Q} \in \mathbb{R}^{d_1 \times \cdots \times d_N}.
\end{equation}
The discretization levels determine the mode sizes $(d_1,\ldots,d_N)$. The total 
number of discretized states is $|\mathcal{S}| = \prod_{n=1}^{D_S} d_n$, which grows 
exponentially with dimension $D_S$. This exponential growth motivates the use of 
low-rank parameterizations that scale as $O(R \sum_n d_n)$ rather than $O(|\mathcal{S}|)$.

The CP (CANDECOMP/PARAFAC) decomposition represents a tensor as a sum of rank-one components, providing a compact parameterization when the tensor admits low-rank structure. The specific formulation and computational details are presented in Section~\ref{subsubsec:tensor_update}, where we describe how TEQL maintains and updates this representation during learning.

In related low-rank reinforcement learning formulations, additional regularity conditions such as coherence or spikiness are often used to formalize when value-function structure can be recovered from limited observations; our analysis here is stated directly in terms of CP approximability.

Having established the representational form and parameterization, we next specify the structural assumption that links this model class to the optimal Q-function.

\subsection{Structural Assumptions for Tensor Value Functions}
\label{subsec:structural_assumptions}

The tensor representation defined above describes a compact model class for approximating 
value functions in high-dimensional MDPs. To ensure that this representation supports 
effective learning, we specify a structural condition on the optimal Q-function $Q^*$.
This assumption constrains how well $Q^*$ can be approximated within the class of 
rank-$R$ CP tensors.

\begin{assumption}[Low-rank CP Approximability]
\label{assump:lowrank}
For a given rank parameter $R$, the optimal value tensor $Q^*$ admits a
rank-$R$ CP approximation up to residual error $B_R$, in the sense that
\begin{equation}
\label{eq:low_rank_approx}    
\inf_{\operatorname{rank}_{\mathrm{CP}}(\mathcal{Q}) \le R}
\left\|\mathcal{Q} - \mathcal{Q}^* \right\|_{\infty}
\;\le\; B_R.
\end{equation}
\end{assumption}

Assumption~\ref{assump:lowrank} does not require that $Q^*$ be exactly
low rank. Rather, it states that $Q^*$ can be well approximated by a
rank-$R$ CP tensor, with the approximation error absorbed into the
constant $B_R$. This is analogous to low-rank approximations widely used
in multivariate analysis, where the dominant structure of a
high-dimensional function can often be captured by a small number of
multiplicative components even though the true function is not strictly
separable. We emphasize that this assumption concerns the value function $Q^*$, not the state-action space itself. The assumption concerns the value function mapping, not the dimensionality of the state or action spaces.

A low-rank approximation is particularly plausible in environments where
the state and action variables interact through a moderate number of
latent or weakly coupled factors. Examples include settings with
approximately separable physical effects, smooth dependencies across
dimensions, or dynamics that evolve on a lower-dimensional manifold
embedded in the ambient space. In such cases, the dominant variation in
$Q^*$ can often be represented with a limited number of rank-one tensor
components, leading to a small $B_R$.

Conversely, when interactions across dimensions are highly entangled or
exhibit strong discontinuities, the approximation error $B_R$ may remain
non-negligible for any feasible rank $R$. This reflects a design trade-off: the approximation error $B_R$ can be systematically reduced by increasing the tensor rank $R$, albeit at the cost of increased computational complexity.

Assumption~\ref{assump:lowrank} requires only that there exist mode sizes
$(d_1,\ldots,d_N)$ and a rank $R$ such that the approximation error
$B_R$ is bounded. The theoretical guarantees in
Section~\ref{subsubsec:theoretical_guarantees} do not impose additional
restrictions on these quantities. In practice, we choose 
$(d_1,\ldots,d_N)$ and $R$ so that the parameter count 
$R \sum_{n=1}^{N} d_n$ is much smaller than the full tensor size 
$\prod_{n=1}^{N} d_n$, preserving the computational and statistical 
advantages of the low-rank representation while allowing any residual
approximation error to be absorbed into $B_R$.

\section{Methodology}\label{sec-meth}

\begin{figure}[H]
\centering
\includegraphics[width=0.6\textwidth]{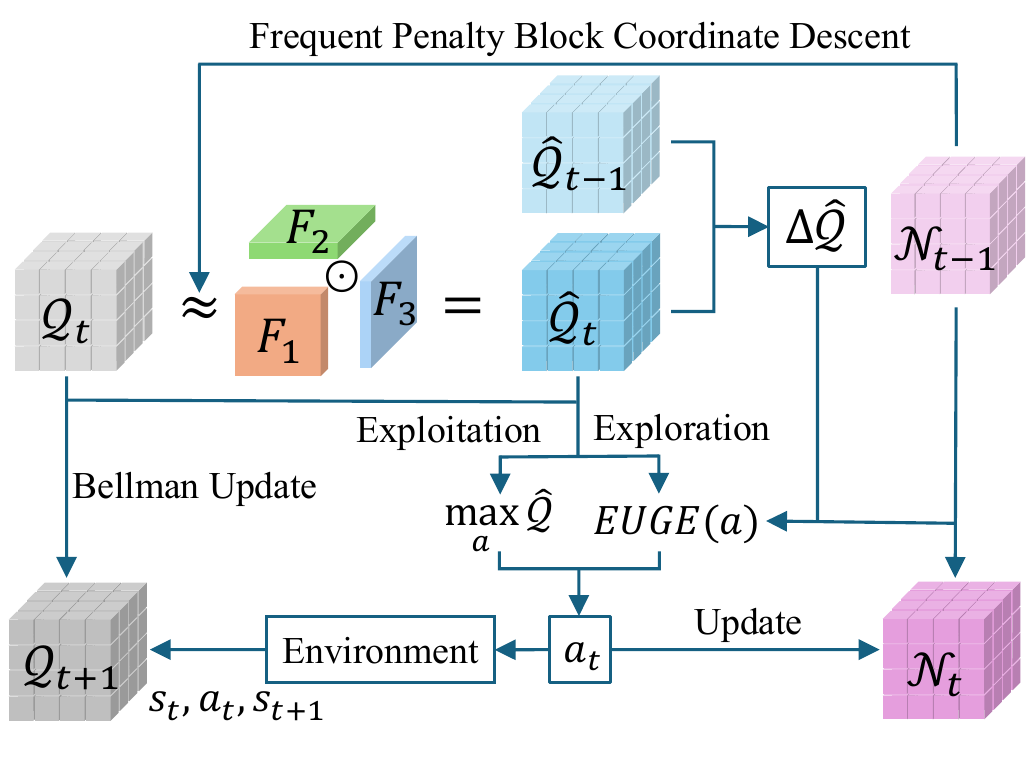}
\caption{\label{fig:flowchart}Framework of the Tensor-Efficient $Q$-Learning (TEQL) Algorithm.}
\end{figure}

Figure~\ref{fig:flowchart} summarizes the Tensor-Efficient $Q$-Learning (TEQL) 
framework. The algorithm maintains a low-rank CP approximation $\hat{\mathcal{Q}}_t$ and iterates through three stages: action selection via EUGE, 
environment interaction, and tensor update via frequency-regularized block coordinate 
descent. The decomposition error $\Delta\mathcal{Q}_t$ and visit counts $\mathcal{N}_t$ 
are carried forward to guide subsequent exploration, forming a closed loop.

At time step $t$, the algorithm starts from the previous estimate $\hat{\mathcal{Q}}_{t-1}$, represented by a rank-$R$ CP decomposition with factor matrices $\{\boldsymbol{F}_n^{(t-1)}\}_{n=1}^N$. Given the current state $s_t$, an action $a_t$ is selected using the Error-Uncertainty Guided Exploration (EUGE) mechanism, which combines the estimated value $\hat{\mathcal{Q}}_{t-1}(s_t,a)$ with an exploration bonus derived from visit counts and approximation changes.

After executing $a_t$, the agent observes a reward $r_t$ and next state $s_{t+1}$, forming a transition tuple $(s_t,a_t,r_t,s_{t+1})$. The tuple and the previous estimate $\hat{\mathcal{Q}}_{t-1}$ are used to update the approximation by solving a frequency-regularized tensor decomposition problem via block coordinate descent, yielding an updated estimate $\hat{\mathcal{Q}}_{t}$.

The change between successive approximations, $\Delta \mathcal{Q}_t(s_t,a_t) = \bigl| \hat{\mathcal{Q}}_{t}(s_t,a_t) - \hat{\mathcal{Q}}_{t-1}(s_t,a_t) \bigr|$, is recorded together with the updated visit counts $\mathcal{N}_t$. These quantities are carried forward to guide subsequent action selection through EUGE. Throughout, the CP rank is denoted by $R$, and the index $r \in \{1,\ldots,R\}$ is used only as a summation index within the decomposition.
\subsection{Low-Rank Tensor Q-Function Update with Frequency Regularization}
\label{subsubsec:tensor_update}

TEQL approximates the Q-function as a low-rank tensor $\hat{\mathcal{Q}}_t \in \mathbb{R}^{d_1 \times \cdots \times d_N}$, where each mode corresponds to one discretized state or action component and $N = D_S + D_A$ is the total number of state and action dimensions. The CP (CANDECOMP/PARAFAC) decomposition represents this tensor as a sum of rank-one components. For an index tuple $(i_1,\ldots,i_N)$, the CP representation is
\[
\hat{\mathcal{Q}}_t(i_1,\ldots,i_N)
=
\sum_{r=1}^{R}
\prod_{n=1}^{N}
\boldsymbol{F}_n^{(t)}(i_n,r),
\]
where $\boldsymbol{F}_n^{(t)} \in \mathbb{R}^{d_n \times R}$ are factor matrices at time $t$. This parameterization reduces the number of parameters from $\prod_{n=1}^N d_n$ in a tabular representation to $R \sum_{n=1}^N d_n$, which we denote by $d_{\mathrm{eff}}$ for subsequent analysis.

TEQL builds on a baseline tensor Q-learning update \citep{rozada2024tensor}, which minimizes squared temporal-difference error at observed pairs. In particular, given $(s_t,a_t,r_t,s_{t+1})$, the baseline corresponds to optimizing a local per-sample objective of the form
\begin{equation}
\min_{\mathcal{Q}}
\left(
q^{\text{target}}_t(s_t,a_t)
-
\mathcal{Q}(s_t,a_t)
\right)^2.
\label{eq:baseline_objective}
\end{equation}

This baseline can put most weight on frequently visited state-action pairs and may allocate fewer updates to rarely visited pairs.

To adjust the allocation of updates, TEQL introduces a frequency-based regularization term, drawing on the principle of visit-count weighting from \citep{auer2002finite,jaksch2010near}. At time $t$ the update solves
\begin{equation}
\begin{split}
\min_{\{\boldsymbol{F}_n^{(t)}\}_{n=1}^N} \Bigl[ \bigl( q^{\text{target}}_t(s_t,a_t) - \hat{\mathcal{Q}}_t(s_t,a_t) \bigr)^2 - \lambda \frac{ \hat{\mathcal{Q}}_t(s_t,a_t)^2 }{ \mathcal{N}_{t-1}(s_t,a_t)+\epsilon } \Bigr],\\
\text{s.t.} \quad \hat{\mathcal{Q}}_t = \sum_{r=1}^R \boldsymbol{F}_1^{(t)}(:,r) \circ \boldsymbol{F}_2^{(t)}(:,r) \circ \cdots \circ \boldsymbol{F}_N^{(t)}(:,r).
\end{split}
\label{eq:target_q}
\end{equation}
where $q^{\text{target}}_t(s_t,a_t) = r_t + \gamma \max_{a'} \hat{\mathcal{Q}}_{t-1}(s_{t+1},a')$ is the target value, $\lambda > 0$ controls the regularization strength, $\epsilon > 0$ avoids division by zero, and $\mathcal{N}_{t-1}(s_t,a_t)$ is the visit count of $(s_t,a_t)$ before time $t$.

We use the squared TD error to form a differentiable objective suitable for gradient-based block coordinate descent. The per-sample loss combining squared TD error and frequency regularization is
\begin{equation}
\boldsymbol{L}_{s_t,a_t}
=
\frac{1}{2}
\Bigl(
q^{\text{target}}_t(s_t,a_t)
-
\hat{\mathcal{Q}}_t(s_t,a_t)
\Bigr)^2
-
\lambda
\frac{
\hat{\mathcal{Q}}_t(s_t,a_t)^2
}{
\mathcal{N}_{t-1}(s_t,a_t)+\epsilon
}.
\label{eq:loss_function}
\end{equation}

The regularization term introduces frequency-dependent regularization. Unlike standard $\ell_2$ regularization that uniformly shrinks values toward zero, this term interacts with the temporal-difference term in a direction-dependent manner. To see this, note that the derivative of~\eqref{eq:loss_function} with respect to $\hat{\mathcal{Q}}_t$ is
\[
\frac{\partial \boldsymbol{L}_{s_t,a_t}}{\partial \hat{\mathcal{Q}}_t}
=
-\bigl(q^{\text{target}}_t-\hat{\mathcal{Q}}_t\bigr)
-
\frac{2\lambda \hat{\mathcal{Q}}_t}{\mathcal{N}_{t-1}+\epsilon},
\]
where we abbreviate $q^{\text{target}}_t = q^{\text{target}}_t(s_t,a_t)$, $\hat{\mathcal{Q}}_t = \hat{\mathcal{Q}}_t(s_t,a_t)$, and $\mathcal{N}_{t-1} = \mathcal{N}_{t-1}(s_t,a_t)$.
When $\hat{\mathcal{Q}}_t$ is positive and exceeds the target (overestimation), the first term is positive and pushes $\hat{\mathcal{Q}}_t$ downward via gradient descent, while the second term is negative and resists this correction. The two terms act in opposite directions, making the update conservative.
When $\hat{\mathcal{Q}}_t$ is below the target (underestimation), both terms are negative and align in the same direction, allowing the correction to proceed unimpeded.
The coefficient $1/(\mathcal{N}_{t-1}+\epsilon)$ modulates the strength of this direction-dependent effect: at frequently visited pairs where $\mathcal{N}_{t-1}$ is large, the regularization term is negligible and TD learning dominates; at rarely visited pairs where $\mathcal{N}_{t-1}$ is small, the regularization effect is pronounced, providing stronger resistance to overestimation while not impeding underestimation corrections.
Since overestimation induced by the maximization operator is a primary source of instability in Q-learning~\citep{thrun1993issues,van2016deep}, this mechanism is particularly valuable at sparsely visited state-action pairs where such bias is most likely to occur.
In summary, at frequently visited pairs, the regularization coefficient is small and TD learning dominates with moderate update magnitudes. At rarely visited pairs, the regularization coefficient is large and the direction-dependent effect becomes significant: overestimation is dampened, while underestimation triggers accelerated correction. This is desirable because rarely visited pairs are precisely where estimates are most uncertain and require larger updates when underestimated.

While this mechanism does not directly promote exploration, it shapes the distribution of $\Delta\mathcal{Q}_t$ and subsequently affects exploration behavior through the EUGE strategy described in Section~\ref{subsec:euge}.

For each observed pair, the decomposition error is defined as
\begin{equation}
\Delta\mathcal{Q}_t(s_t,a_t)
=
\bigl|
\hat{\mathcal{Q}}_{t}(s_t,a_t)
-
\hat{\mathcal{Q}}_{t-1}(s_t,a_t)
\bigr|.
\label{eq:decomp_error}
\end{equation}
This quantity is distinct from the TD error and reflects the change induced by the tensor update. EUGE uses this decomposition error to construct an exploration bonus.

The gradient of~\eqref{eq:loss_function} with respect to a factor entry $\boldsymbol{F}_n(i_n,r)$ is
\begin{equation}
\begin{split}
\nabla_{\boldsymbol{F}_n(i_n,r)} \boldsymbol{L}_{s_t,a_t}
&=
-
\Bigl(
q^{\text{target}}_t(s_t,a_t)
-
\hat{\mathcal{Q}}_t(s_t,a_t)
\Bigr)
\prod_{m \neq n} \boldsymbol{F}_m(i_m,r) \\
&\quad
-
2\lambda
\frac{
\hat{\mathcal{Q}}_t(s_t,a_t)
}{
\mathcal{N}_{t-1}(s_t,a_t)+\epsilon
}
\prod_{m \neq n} \boldsymbol{F}_m(i_m,r).
\end{split}
\label{eq:gradient}
\end{equation}
During this computation, $q^{\text{target}}_t(s_t,a_t)$ is treated as a constant because it is defined using $\hat{\mathcal{Q}}_{t-1}$ from the previous step.
Only $\hat{\mathcal{Q}}_t(s_t,a_t)$ depends on the current factor matrices.

\begin{algorithm}[!htbp]
\footnotesize
\caption{Low-Rank Tensor Q-Function Update}
\label{alg:update}
\begin{algorithmic}[1]
\State \textbf{Input}: State $s_t$, action $a_t$, reward $r_t$, next state $s_{t+1}$, Q-function $\hat{\mathcal{Q}}_{t-1}$, factor matrices $\boldsymbol{F}_n^{(t-1)}$, visit count $\mathcal{N}_{t-1}(s_t,a_t)$, error tensor $\Delta\mathcal{Q}_{t-1}$, regularization parameter $\lambda$, regularization term constant $\epsilon$, learning rate parameters $\alpha_0,\kappa$, time step $t$, threshold $\tau$, maximum inner iterations $I_{\max}$.
\State Set step size $\alpha_t = \alpha_0/(1+\kappa t)$.
\State Compute target value
$q^{\text{target}}_t(s_t,a_t) = r_t + \gamma \max_{a'} \hat{\mathcal{Q}}_{t-1}(s_{t+1},a')$.
\Comment{$q^{\text{target}}_t$ depends only on $\hat{\mathcal{Q}}_{t-1}$ and is fixed during the inner updates.}
\For{$n = 1,\ldots,N$}
    \State Fix $\boldsymbol{F}_m^{(t-1)}$ for all $m \neq n$.
    \State Set $\hat{\mathcal{Q}}_{\text{prev}} = \hat{\mathcal{Q}}_{t-1}(s_t,a_t)$.
    \For{$i = 1$ to $I_{\max}$}
        \State Define the local loss at $(s_t,a_t)$:
        $
        L_{s_t,a_t}
        =
        \frac{1}{2}\big(q^{\text{target}}_t(s_t,a_t) - \hat{\mathcal{Q}}_t(s_t,a_t)\big)^2
        -
        \lambda \frac{\hat{\mathcal{Q}}_t(s_t,a_t)^2}{\mathcal{N}_{t-1}(s_t,a_t)+\epsilon},
        $
        \Statex \qquad\qquad where $q^{\text{target}}_t$ and $\mathcal{N}_{t-1}$ are treated as constants when updating $\boldsymbol{F}_n$, while $\hat{\mathcal{Q}}_t$ changes through $\boldsymbol{F}_n$.
        \State Compute gradient
        $\nabla_{\boldsymbol{F}_n(i_n,r)} L_{s_t,a_t}$
        as in \eqref{eq:gradient}.
        \State Update
        $\boldsymbol{F}_n(i_n,r)
        \leftarrow
        \boldsymbol{F}_n(i_n,r)
        -
        \alpha_t
        \nabla_{\boldsymbol{F}_n(i_n,r)} L_{s_t,a_t}$.
        \State Recompute
        $\hat{\mathcal{Q}}_{\text{curr}} = \hat{\mathcal{Q}}_t(s_t,a_t)$.
        \If{$|\hat{\mathcal{Q}}_{\text{curr}} - \hat{\mathcal{Q}}_{\text{prev}}| < \tau$}
            \State \textbf{break}
        \EndIf
        \State $\hat{\mathcal{Q}}_{\text{prev}} = \hat{\mathcal{Q}}_{\text{curr}}$.
    \EndFor
\EndFor
\State Set $\boldsymbol{F}_n^{(t)} = \boldsymbol{F}_n$.
\State Update
$\hat{\mathcal{Q}}_t(s,a) = \sum_{r=1}^{R} \prod_{n=1}^{N} \boldsymbol{F}_n^{(t)}(i_n,r)$.
\State Set
$\Delta\mathcal{Q}_t(s_t,a_t)
=
|\hat{\mathcal{Q}}_{t}(s_t,a_t) - \hat{\mathcal{Q}}_{t-1}(s_t,a_t)|$.
\State Update visit count $\mathcal{N}_t(s_t,a_t) = \mathcal{N}_{t-1}(s_t,a_t)+1$.
\State \textbf{Output}: $\hat{\mathcal{Q}}_t$, $\boldsymbol{F}_n^{(t)}$, $\Delta\mathcal{Q}_t$, $\mathcal{N}_t$.
\end{algorithmic}
\end{algorithm}

At each time step $t$, the factor matrices $\boldsymbol{F}_n^{(t-1)}$ are used to initialize the current optimization variables, denoted by $\boldsymbol{F}_n$, which are updated in-place during the inner block coordinate descent and recorded as $\boldsymbol{F}_n^{(t)}$ upon completion. The factor matrices $\boldsymbol{F}_n$ are updated by gradient descent:
\begin{equation}
\boldsymbol{F}_n(i_n,r)
\leftarrow
\boldsymbol{F}_n(i_n,r)
-
\alpha_t
\nabla_{\boldsymbol{F}_n(i_n,r)} \boldsymbol{L}_{s_t,a_t},
\label{eq:factor_update}
\end{equation}
where $\alpha_t$ is a step size. Updates proceed until the change in $\hat{\mathcal{Q}}_t(s_t,a_t)$ is below a threshold or a maximum number of iterations is reached.

Algorithm~\ref{alg:update} summarizes the low-rank tensor Q-function update.
\subsection{Error-Uncertainty Guided Exploration (EUGE)}
\label{subsec:euge}

The tensor update produces $\Delta\mathcal{Q}_t$, which quantifies estimation uncertainty beyond visit counts alone. EUGE leverages this signal for action selection: At time $t$, for each action $a$ in state $s_t$, EUGE defines
\begin{equation}
\text{EU}_t(s_t,a)
=
\hat{\mathcal{Q}}_{t-1}(s_t,a)
+
c
\left(
\Delta\mathcal{Q}_{t-1}(s_t,a)
+
\sqrt{
\frac{
\log \mathcal{N}_{\text{total},t-1}(s_t)
}{
\mathcal{N}_{t-1}(s_t,a) + 1
}
}
\right),
\label{eq:euge}
\end{equation}
where $\mathcal{N}_{t-1}(s_t,a)$ is the visit count of $(s_t,a)$, $\mathcal{N}_{\text{total},t-1}(s_t) = \sum_a \mathcal{N}_{t-1}(s_t,a)$ is the total visits to $s_t$, and $c>0$ is an exploration parameter. The action is chosen as
\[
a_t = \arg\max_a \text{EU}_t(s_t,a).
\]
The EUGE value combines the current tensor estimate $\hat{\mathcal{Q}}_{t-1}(s_t,a)$, the decomposition error $\Delta\mathcal{Q}_{t-1}(s_t,a)$, and a visit-count-based bonus. Larger decomposition error or smaller visit count yields a larger bonus, prioritizing actions whose Q-value estimates are either changing rapidly or have been sampled infrequently. Algorithm~\ref{alg:euge} summarizes EUGE. The form of the bonus follows the standard UCB principle of favoring uncertain actions, with the additional decomposition-error term reflecting uncertainty induced by low-rank approximation rather than visitation alone.

\begin{algorithm}
\caption{Error-Uncertainty Guided Exploration (EUGE)}
\label{alg:euge}
\begin{algorithmic}[1]
\State \textbf{Input}: State $s_t$, tensor Q-function $\hat{\mathcal{Q}}_{t-1}$, factor matrices $\boldsymbol{F}_n^{(t-1)}$, error tensor $\Delta\mathcal{Q}_{t-1}$, visit counts $\mathcal{N}_{t-1}$, parameter $c$.
\State Initialize possible actions $\mathcal{A}_{\text{possible}}$.
\For{each $a \in \mathcal{A}_{\text{possible}}$}
    \State Compute $\hat{\mathcal{Q}}_{t-1}(s_t,a)$ using $\boldsymbol{F}_n^{(t-1)}$.
    \State Compute
    $\text{bonus}_t(s_t,a)
    =
    \Delta\mathcal{Q}_{t-1}(s_t,a)
    +
    \sqrt{
    \frac{
    \log \mathcal{N}_{\text{total},t-1}(s_t)
    }{
    \mathcal{N}_{t-1}(s_t,a)+1
    }
    }$.
    \State Set
    $\text{EU}_t(s_t,a)
    =
    \hat{\mathcal{Q}}_{t-1}(s_t,a)
    +
    c \cdot \text{bonus}_t(s_t,a)$.
\EndFor
\State Select $a_t = \arg\max_a \text{EU}_t(s_t,a)$.
\State \textbf{Output}: $a_t$.
\end{algorithmic}
\end{algorithm}

\subsection{Complete TEQL Algorithm and Model Configuration}
\label{subsec:teql_algorithm}
 
The TEQL framework integrates the low-rank tensor update (Algorithm~\ref{alg:update}) and the EUGE exploration strategy (Algorithm~\ref{alg:euge}) into an online learning loop. At each time step, the agent uses EUGE to select an action based on the current Q-function estimate and uncertainty measures, observes the resulting transition, and then updates the tensor factors via block coordinate descent. The decomposition error $\Delta\mathcal{Q}_t$ computed during the update step feeds back into the EUGE bonus for subsequent action selection, creating a coupling between representation learning and exploration.
 
TEQL operates in an infinite-horizon discounted Markov Decision Process defined by $\langle \mathcal{S}, \mathcal{A}, \mathcal{P}, \mathcal{R}, \gamma \rangle$, where $\mathcal{S}$ and $\mathcal{A}$ are the state and action spaces, $\mathcal{P}$ is the transition kernel, $\mathcal{R}$ is the reward function, and $\gamma \in (0, 1)$ is the discount factor. The algorithm runs for $T$ episodes, each consisting of $H$ steps. Algorithm~\ref{alg:teql} presents the complete procedure. The computational complexity of TEQL's tensor update via block coordinate descent is $\mathcal{O}(N R I_{\max})$ per update, where $N$ is the tensor order, $R$ the rank, and $I_{\max}$ the maximum inner iterations. Memory usage is $\mathcal{O}(d_{\mathrm{eff}})$, as defined in Section~\ref{subsubsec:tensor_update}.
 
\begin{algorithm}[!htbp]
\caption{Tensor-Efficient Q-Learning (TEQL)}
\label{alg:teql}
\begin{algorithmic}[1]
\State \textbf{Input}: Discount factor $\gamma$, tensor rank $R$, regularization parameter $\lambda$, regularization constant $\epsilon$, learning rate parameters $\alpha_0, \kappa$, exploration constant $c$, convergence threshold $\tau$, maximum inner iterations $I_{\max}$, number of episodes $T$, episode length $H$.
\State Initialize factor matrices $\boldsymbol{F}_n^{(0)} \in \mathbb{R}^{d_n \times R}$ for $n = 1, \ldots, N$.
\State Initialize Q-function $\hat{\mathcal{Q}}_0(s,a) = \sum_{r=1}^{R} \prod_{n=1}^{N} \boldsymbol{F}_n^{(0)}(i_n, r)$.
\State Initialize visit counts $\mathcal{N}_0(s,a) = 0$ for all $(s,a)$.
\State Initialize error tensor $\Delta\mathcal{Q}_0(s,a) = 0$ for all $(s,a)$.
\For{episode $e = 1, \ldots, T$}
    \State Observe initial state $s_1$ from environment.
    \For{step $h = 1, \ldots, H$}
        \State Select action $a_h$ using EUGE (Algorithm~\ref{alg:euge}) with inputs $(s_h, \hat{\mathcal{Q}}, \boldsymbol{F}_n, \Delta\mathcal{Q}, \mathcal{N}, c)$.
        \State Execute $a_h$, observe reward $r_h$ and next state $s_{h+1}$.
        \State Update $(\hat{\mathcal{Q}}, \boldsymbol{F}_n, \Delta\mathcal{Q}, \mathcal{N})$ using Low-Rank Tensor Update (Algorithm~\ref{alg:update}) with inputs $(s_h, a_h, r_h, s_{h+1}, \hat{\mathcal{Q}}, \boldsymbol{F}_n, \mathcal{N}, \Delta\mathcal{Q}, \lambda, \epsilon, \alpha_0, \kappa, t, \tau, I_{\max})$, where $t = (e-1)H + h$.
    \EndFor
\EndFor
\State \textbf{Output}: Learned Q-function $\hat{\mathcal{Q}}$, factor matrices $\boldsymbol{F}_n$.
\end{algorithmic}
\end{algorithm}
 
Algorithm~\ref{alg:teql} involves several quantities that control the learning process. We organize them into three categories and provide general selection guidelines below; the specific values used in our experiments are reported in Section~\ref{subsec:hyperparameters}.
 
\textbf{Model-capacity parameters.}  The discretization levels $(d_1,\ldots,d_N)$ and the tensor rank $R$ must be specified before running TEQL. These quantities determine the tensor shape of $\hat{\mathcal{Q}}$ and the parameter count $R\sum_{n=1}^N d_n$, which governs the statistical and computational complexity of estimating the tensor factors and appears explicitly in the error bounds in Section~\ref{subsubsec:theoretical_guarantees}. TEQL does not adaptively modify $(d_1,\ldots,d_N)$ or $R$ during learning; both are fixed hyperparameters chosen prior to training. When state or action variables arise from continuous domains, each dimension is mapped to a finite set of indices through uniform binning. The number of bins $d_n$ is selected so that (i) the discretization covers the full admissible range of that dimension, and (ii) the resulting discrete state-action space remains compatible with available computation. This procedure does not require environment-specific domain knowledge; the same uniform binning scheme applies across different tasks. The tensor rank $R$ is chosen to satisfy $R\sum_{n=1}^N d_n \ll \prod_{n=1}^N d_n$, so that the low-rank representation yields a substantial reduction in the number of parameters relative to a full tabular representation. Because the intrinsic CP rank of $Q^*$ and the approximation error $B_R$ are generally unknown, $R$ is treated as a tunable parameter. We select $R$ following prior work on low-rank value-function approximation, where moderate ranks are commonly used to balance approximation accuracy and parameter efficiency. In the experimental section, we report sensitivity results across different discretization levels while keeping $R$ fixed, in order to isolate the effect of discretization and assess robustness. This configuration procedure specifies a complete and fixed tensor model class prior to training. The structural assumption in Section~\ref{subsec:structural_assumptions} determines whether this model class admits a sufficiently accurate approximation of $Q^*$. If the approximation error $B_R$ is large, this limitation appears as a non-negligible approximation bias in the neighbourhood radius of Theorem~\ref{thm:convergence} and in the finite-sample error behaviour discussed in Section~\ref{subsubsec:theoretical_guarantees}.
 
\textbf{Numerical optimization parameters.} Four parameters govern the optimization dynamics of the block coordinate descent. The initial learning rate $\alpha_0$ controls the step size of Q-value updates; values that are too large cause divergence, while values that are too small slow convergence. The decay coefficient $\kappa$ in the schedule $\alpha_t = \alpha_0/(1+\kappa t)$ ensures asymptotic convergence; standard choices from the stochastic approximation literature apply. The convergence threshold $\tau$ for inner iterations terminates the block coordinate descent when the change in $\hat{\mathcal{Q}}_t(s_t,a_t)$ falls below $\tau$; in practice, early termination typically occurs within 5 to 10 iterations. An optional parameter $I_{\max}$ can be used to cap the number of inner iterations when computational cost is a concern. The smoothing constant $\epsilon$ in the regularization term prevents division by zero when visit counts are small; any value that is negligible relative to typical visit counts suffices. Within a broad range, moderate changes to these parameters affect convergence speed but not asymptotic performance.
 
\textbf{Algorithm-specific hyperparameters.} Two hyperparameters directly influence TEQL's exploration and regularization behavior. The frequency regularization strength $\lambda$ controls the intensity of the visit-count-dependent regularization in~\eqref{eq:target_q}. Larger $\lambda$ provides stronger resistance to value changes at sparsely visited pairs, counteracting overestimation but potentially slowing convergence. In practice, $\lambda$ should be small enough to preserve TD learning dynamics while providing measurable stabilization. The exploration coefficient $c$ in the EUGE bonus~\eqref{eq:euge} scales the exploration bonus. This parameter should be calibrated so that the exploration bonus is commensurate with the range of Q-values in the problem; if typical values lie in $[0, V_{\max}]$, then $c$ of order $O(1)$ to $O(V_{\max})$ is appropriate. Both $\lambda$ and $c$ can be calibrated on a single representative environment using early-stage learning curves and then held fixed across all experiments. Section~\ref{subsec:sensitivity} reports sensitivity analysis demonstrating robustness within a broad range around the chosen values.
\subsection{Theoretical Guarantees}
\label{subsubsec:theoretical_guarantees}

This section provides theoretical support for TEQL that formalizes the design choices in Sections~\ref{sec-prelim}-\ref{sec-meth}. Rather than aiming for exact optimality guarantees, our goal is to understand how approximation, regularization, and exploration interact to produce stable and efficient learning behavior in high-dimensional settings.

Assumption~\ref{assump:lowrank} (Low-rank CP Approximability) has already been 
introduced in Section~\ref{subsec:structural_assumptions}. 
%
It characterizes the expressive power of the rank-$R$ tensor class and captures 
the irreducible modeling bias through the constant $B_R$.
Here we impose only a 
standard boundedness condition, which can be enforced in practice by factor 
clipping or projection in the CP parameterization.

\begin{assumption}[Bounded Iterates]
\label{assump:bounded_iterates}
The TEQL iterates remain uniformly bounded: 
$\|\hat{\mathcal{Q}}_t\|_{\infty} \le V_{\max}$ for all $t \ge 0$.
\end{assumption}

This assumption ensures that all iterates stay within a compact region where 
the approximation error characterized by Assumption~\ref{assump:lowrank} 
remains meaningful and where the Bellman operator is well behaved.

We state the convergence result in interpretable form; the detailed recursion 
is deferred to the appendix.

\begin{theorem}[Convergence to a Neighborhood]
\label{thm:convergence}
Under Assumptions~\ref{assump:lowrank} and~\ref{assump:bounded_iterates}, TEQL 
converges to a neighborhood of the optimal Q-function $q^*$ in expected supremum 
norm. The asymptotic error decomposes into two irreducible components: 
(i) an approximation bias determined by $B_R$ from Assumption~\ref{assump:lowrank}, 
and (ii) a stochastic component induced by using the single-sample TD 
target~\eqref{eq:target_q}. 
\end{theorem}

A finite-time bound with explicit error decomposition is given in Appendix~\ref{app:convergence_proof}. Theorem~\ref{thm:convergence} shows that TEQL behaves like a contractive 
Bellman-type iteration (due to $\gamma \in (0,1)$), but cannot converge exactly 
to $q^*$ unless (i) the rank-$R$ class is expressive enough ($B_R$ small), and 
(ii) TD noise vanishes. 
%
This result clarifies the fundamental performance limit of TEQL: stability and 
accuracy are achieved up to unavoidable approximation and sampling effects, 
rather than through exact Bellman fixed-point recovery.
In other words, TEQL is stable and accurate up to the 
modeling error $B_R$ and the inherent sampling noise.

Having established that TEQL converges in a controlled manner, we next examine 
how the algorithm allocates its updates across the state-action space.

The regularizer in~\eqref{eq:loss_function} introduces a visit-count-dependent term in the gradient. Proposition~\ref{prop:freq_shrinkage} establishes an upper bound on the update magnitude $|\Delta\mathcal{Q}_t(s_t,a_t)|$ that includes a component scaling as $1/(\mathcal{N}_{t-1}(s_t,a_t)+\epsilon)$. This bound is tighter at frequently visited pairs, reflecting that the regularization effect diminishes with increasing visit count.

\begin{proposition}[Frequency-Regularized Update Shrinkage]
\label{prop:freq_shrinkage}
Under Assumption~\ref{assump:bounded_iterates}, the magnitude of the one-step 
value change $\Delta\mathcal{Q}_t(s_t,a_t)$ admits an upper bound consisting of 
a base term (depending on the step size and boundedness constants) plus an 
additional term that decays on the order of 
$1/(\mathcal{N}_{t-1}(s_t,a_t)+\epsilon)$. Consequently, for a fixed step-size 
schedule, updates at frequently visited pairs become progressively smaller. 
\end{proposition}

The detailed proof is given in Appendix~\ref{app:freq_shrinkage_proof}.This result formalizes how the upper bound on update magnitude varies with visitation frequency. At frequently visited pairs, the bound is tighter, reflecting that TD learning dominates with minimal regularization interference. At rarely visited pairs, the bound is looser, but as shown in Section~\ref{subsubsec:tensor_update}, the direction-dependent nature of the regularization selectively dampens overestimation while allowing underestimation corrections.

We now analyze the selection rule~\eqref{eq:euge}. 
%
Because EUGE explicitly depends on both visit counts and decomposition error, 
the update shrinkage property in Proposition~\ref{prop:freq_shrinkage} plays a 
direct role in the exploration dynamics.
Importantly, we do not 
introduce any new exploration score: we use exactly $\mathrm{EU}_t(s,a)$ as 
defined in Section~\ref{subsec:euge}. The main takeaway is that EUGE does not 
keep choosing actions that are simultaneously (i) clearly worse than the best 
action at the current state and (ii) already low-uncertainty according to the 
bonus terms in~\eqref{eq:euge}.

\begin{proposition}[Logarithmic Re-Selection of Suboptimal Low-Uncertainty Actions]
\label{prop:euge_selection}
Fix a state $s$. Consider an action $a$ such that, after some time, its 
estimated value $\hat{\mathcal{Q}}_{t-1}(s,a)$ remains separated below the best 
estimated action at $s$ by a fixed positive margin, while its EUGE bonus 
in~\eqref{eq:euge} becomes small due to (i) small decomposition error 
$\Delta\mathcal{Q}_{t-1}(s,a)$ and (ii) a growing visit count 
$\mathcal{N}_{t-1}(s,a)$. Then EUGE selects $a$ only a logarithmic number of 
times as the horizon increases. 
\end{proposition}

\medskip
The proof is provided in 
Appendix~\ref{app:euge_selection_proof}.
Together, Theorem~\ref{thm:convergence} and 
Propositions~\ref{prop:freq_shrinkage}-\ref{prop:euge_selection} provide a 
coherent explanation of TEQL: 
%
convergence ensures that learning stabilizes near an optimal solution, 
frequency regularization governs how updates are distributed across the 
state-action space, and EUGE leverages both effects to guide exploration.
(i) stable learning up to approximation bias 
$B_R$ and TD noise, 
(ii) computational effort automatically reallocated away 
from frequent pairs by frequency regularization, and 
(iii) exploration guided by $\Delta\mathcal{Q}_t$ and visit counts avoids 
repeatedly sampling actions that are both unpromising and already well-estimated.
\section{Numerical Study}
\label{sec:experiments}
This section tests whether tensor low-rank structure improves sample efficiency when all methods are constrained to the same parameter budget. We evaluate on three environments with increasing state-action complexity, namely Pendulum ($D_S=2, D_A=1$), CartPole ($D_S=4, D_A=1$), and Highway ($D_S=9, D_A=1$). These environments are selected because their $Q$-functions admit accurate low-rank approximations; singular value analysis in \citet{rozada2024tensor} shows that rank-10 (Pendulum, Cartpole) and rank-20 (Highway) tensors capture over 90\% of spectral energy under comparable discretization schemes.

To isolate the effect of algorithmic design from model capacity, all methods operate on identical discretized MDPs and are matched to the same order of magnitude in trainable parameters. This parameter-matched design is central to our evaluation because tensor methods are specifically intended for learning in regimes where model capacity is limited. We constrain all methods to equivalent budgets, specifically 500 for Pendulum, 700 for CartPole ($R=10$), and 3,700 for Highway ($R=20$), to directly test whether tensor low-rank structures provide a meaningful advantage. These budgets are deliberately set below typical deep learning scales to focus the evaluation on the intrinsic parameter efficiency of each architecture.

Baselines include (i)~TLR \citep{rozada2024tensor}, which shares TEQL's CP 
representation but uses $\varepsilon$-greedy exploration without regularization, 
(ii)~LoRa-VI, adapted from the matrix low-rank framework of \citet{stojanovic2024model} 
to the Q-learning setting, and (iii)~DQN \citep{mnih2015human} and discrete 
SAC \citep{christodoulou2019sacdiscrete}. We include LoRa-VI not as a direct competitor but to illustrate the limitations of matrix-based low-rank methods in online trajectory-based learning: the original framework assumes uniform or leverage-score-guided sampling for matrix completion, whereas online RL produces correlated, non-uniform data that violates these assumptions.
All methods receive discretized bucket indices as input and output $Q$-values over the finite action set. DQN and SAC are compressed to the same parameter scale by reducing hidden layer widths; SAC further distributes its budget across actor, critic, and temperature networks, leaving each component with fewer parameters than a single DQN network of equivalent total size. Detailed architectures for all methods are provided in Appendix~\ref{appendix:parameter_matching}.

\subsection{Hyperparameter Configuration}
\label{subsec:hyperparameters}
 
Table~\ref{tab:hyperparams} reports the complete hyperparameter configuration for each environment. The discretization column lists the number of bins per dimension in the order they appear in the state-action vector; for example, CartPole has five dimensions discretized into 10, 10, 20, 20, and 10 bins respectively, and Highway has nine state dimensions each with 20 bins and one action dimension with 5 bins. The numerical optimization parameters ($\alpha_0$, $\kappa$, $\tau$, $\epsilon$) were set following standard heuristics from the Q-learning and tensor factorization literature. The algorithm-specific hyperparameters ($\lambda$, $c$) were calibrated on CartPole using early-stage learning curves and then held fixed across all environments. The sensitivity analysis in Section~\ref{subsec:sensitivity} demonstrates robustness within a broad range around these values. Results aggregate 100 independent runs, and shaded regions show mean $\pm$ standard deviation.
 
\begin{table}[h]
\centering
\small
\caption{Hyperparameter settings for all environments.}\label{tab:hyperparams}

\begin{tabular}{llccc}
\toprule
Category & Parameter & Pendulum & CartPole & Highway \\
\midrule
\multirow{2}{*}{Model capacity}
  & Tensor rank $R$ & 10 & 10 & 20 \\
  & Bins per dimension $(d_1,\ldots,d_N)$ & (20, 20, 10) & (10, 10, 20, 20, 10) & (20\,$\times$\,9, 5) \\
\midrule
\multirow{4}{*}{Optimization}
  & Initial learning rate $\alpha_0$ & 0.005 & 0.005 & 0.0002 \\
  & Decay coefficient $\kappa$ & 0.001 & 0.001 & 0.001 \\
  & Inner convergence $\tau$ & 0.01 & 0.01 & 0.01 \\
  & Smoothing constant $\epsilon$ & $10^{-4}$ & $10^{-4}$ & $10^{-4}$ \\
\midrule
\multirow{2}{*}{Algorithm-specific}
  & Regularization $\lambda$ & $10^{-3}$ & $10^{-3}$ & $10^{-3}$ \\
  & Exploration coefficient $c$ & 1.0 & 2.0 & 2.0 \\
\midrule
\multirow{3}{*}{Training}
  & Episodes & 40{,}000 & 10{,}000 & 10{,}000 \\
  & Steps per episode & 100 & 100 & 50 \\
  & Discount factor $\gamma$ & 0.99 & 0.99 & 0.99 \\
\bottomrule
\end{tabular}
\end{table}

\subsection{Classic Control Environments}
\label{subsec:baseline_comparison}

\begin{figure}[h]
\centering
\includegraphics[width=0.9\textwidth]{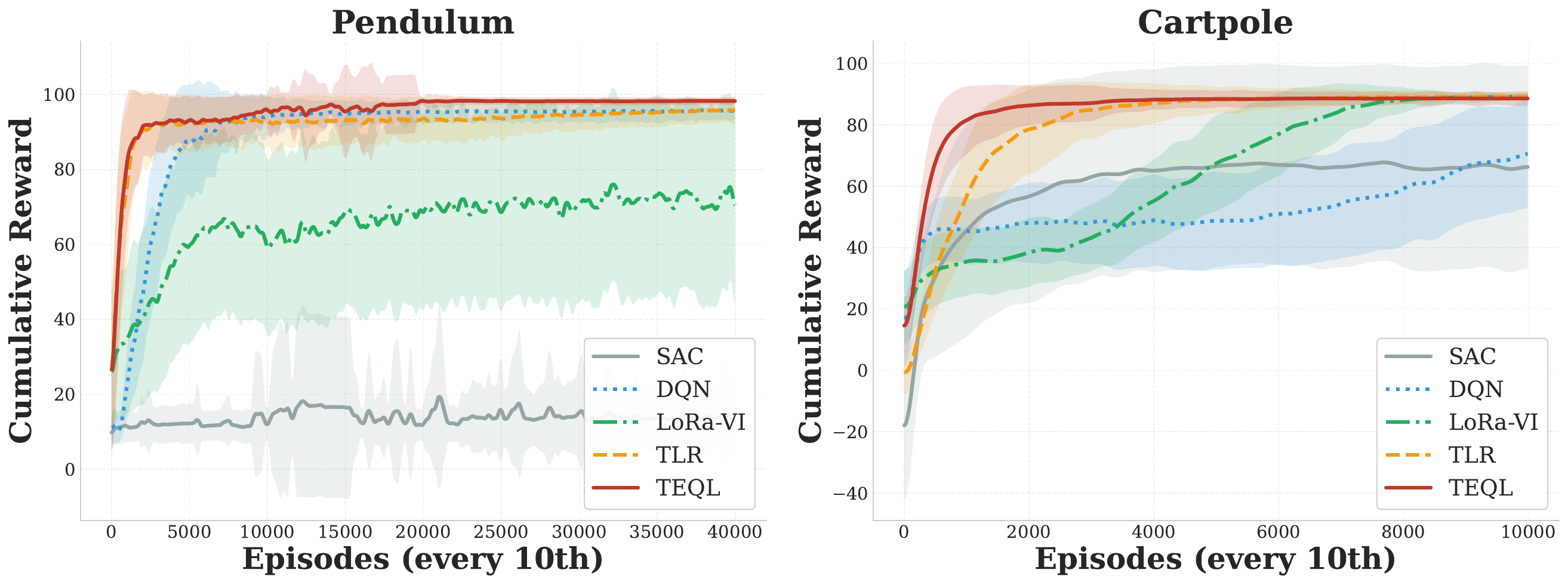} 
\caption{Learning curves on Pendulum (left) and CartPole (right). CP tensor methods (TEQL, TLR) outperform matrix-based (LoRa-VI) and neural baselines (DQN, SAC) under matched parameter budgets. The TEQL-TLR gap widens from Pendulum to CartPole as EUGE's benefit increases with dimensionality. Shaded regions show mean $\pm$ standard deviation over 100 runs.}
\label{fig:pendulum_cartpole}
\end{figure}
Tensor-based methods outperform matrix-based and neural baselines 
under matched parameter budgets. In Figure~\ref{fig:pendulum_cartpole}, LoRa-VI 
learns more slowly because CUR decomposition incurs higher parameter costs than 
CP factorization: CUR stores $K$ anchor rows and columns, requiring 
$O(K(|\mathcal{S}|+|\mathcal{A}|-K))$ parameters. To match the budget, LoRa-VI 
uses only 3 buckets per state dimension in CartPole versus 10-20 for TEQL/TLR, 
which increases approximation bias.

Neural baselines do not merely exhibit slower learning but suffer from representational collapse under these extreme budgets. As the state-space expands, the fixed-capacity MLP fails to resolve the value landscape, whereas the tensor structure maintains a coherent global approximation by leveraging its intrinsic inductive bias rather than raw parameter count.

The advantage of TEQL over TLR scales with dimensionality. In 
Pendulum, the small state-action space allows uniform exploration to achieve 
reasonable coverage. In CartPole, the performance gap widens as undirected 
exploration becomes costly.

\subsection{High-Dimensional Setting}
\label{subsec:highdim}

\begin{figure}[h]
\centering
\includegraphics[width=0.6\textwidth]{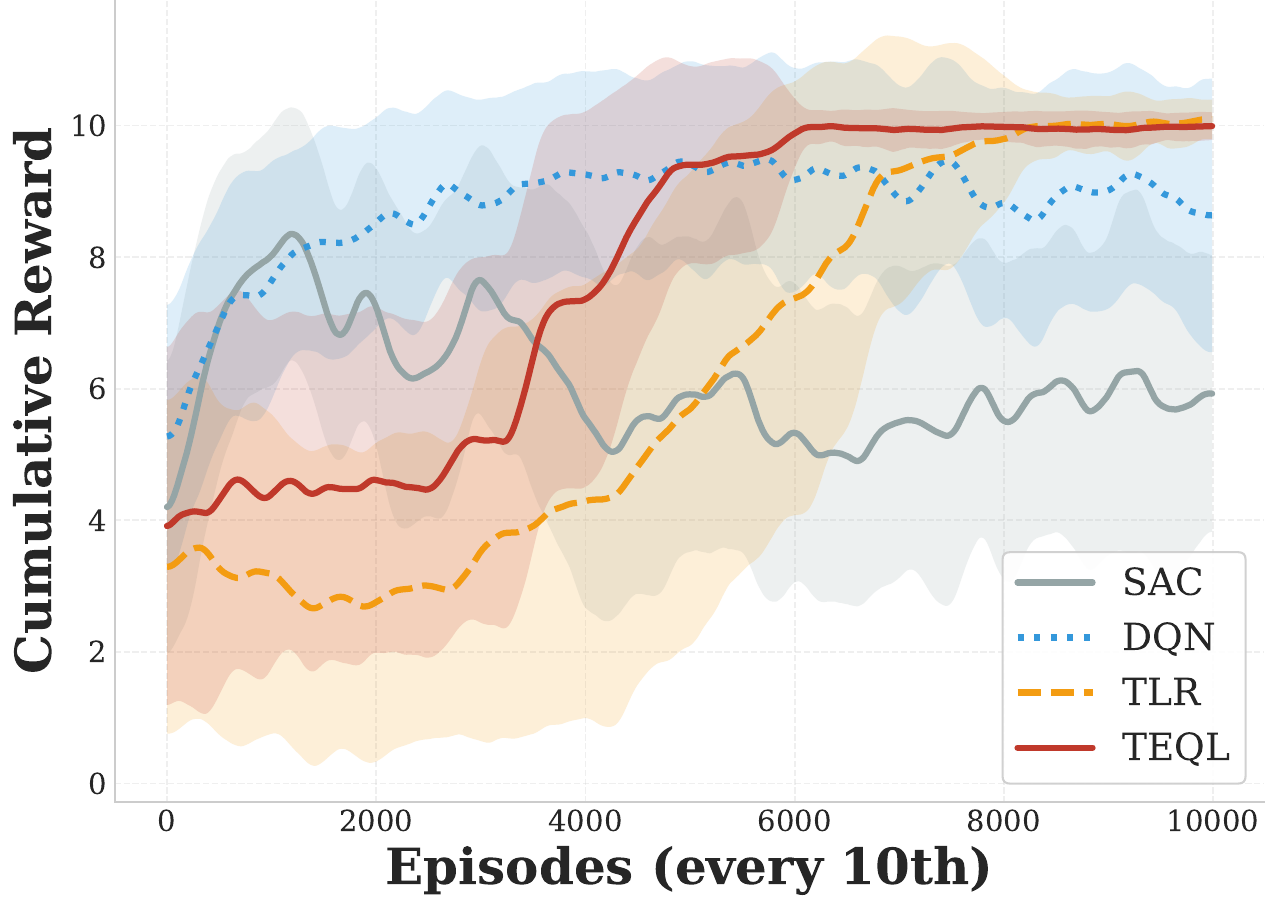} 
\caption{TEQL reaches high performance around episode 5,000, approximately 2,000 episodes before TLR. DQN oscillates; SAC regresses after initial progress. Highway: $D_S{=}9$, $D_A{=}1$.}
\label{fig:highway}
\end{figure}
The Highway environment simulates autonomous driving where an ego vehicle 
navigates multi-lane traffic. The nine-dimensional state space and 
safety-critical nature of the task make it a challenging testbed for 
evaluating scalability beyond the low-dimensional control tasks above.

As shown in Figure~\ref{fig:highway}, TEQL reaches high performance around episode 5,000, approximately 2,000 episodes before TLR. DQN and SAC exhibit persistent instability, failing to retain high-reward policies once discovered. This stability gap in Highway underscores the coupling effect detailed in Section~\ref{subsec:ablation}: in high-dimensional manifolds, local updates in CP factors have global footprints, making frequency-aware regularization a prerequisite for structural integrity. LoRa-VI is excluded because CUR requires $O(|\mathcal{S}|)$ parameters; additionally, narrow trajectory coverage makes leverage score estimation unreliable.

\subsection{Ablation Study: Effect of Regularization Parameter \texorpdfstring{$\lambda$}{lambda}}
\label{subsec:ablation}

\begin{figure}[h]
    \centering
    \includegraphics[width=0.9\textwidth]{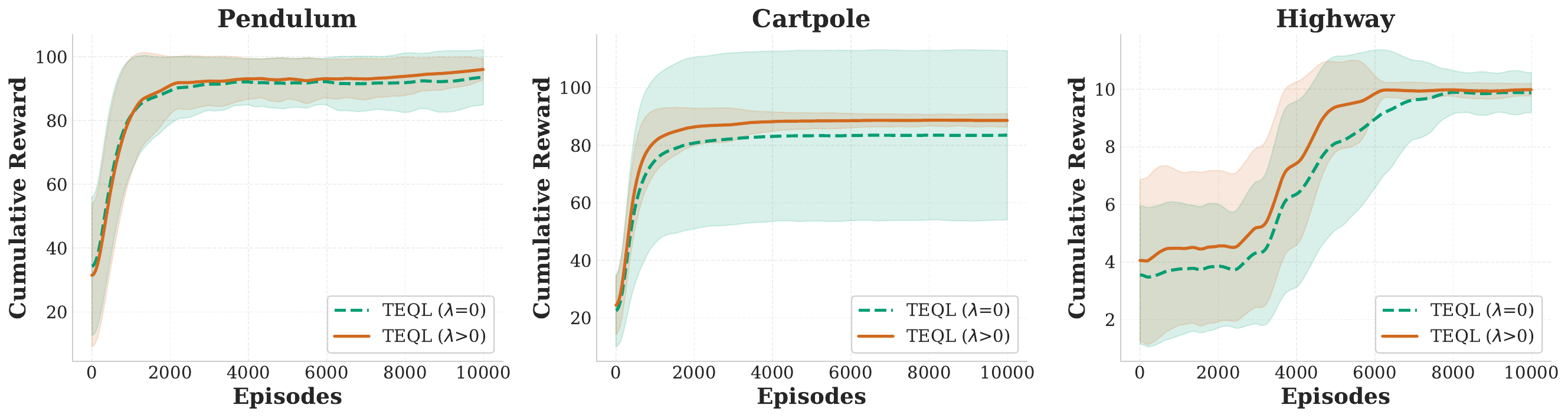}
    \caption{Learning curves for TEQL with $\lambda=0$ and $\lambda>0$. Setting $\lambda>0$ reduces variance and achieves faster convergence across all environments.}
    \label{fig:ablation_curve}
\end{figure}

\begin{figure}[h]
    \centering
    \includegraphics[width=0.9\textwidth]{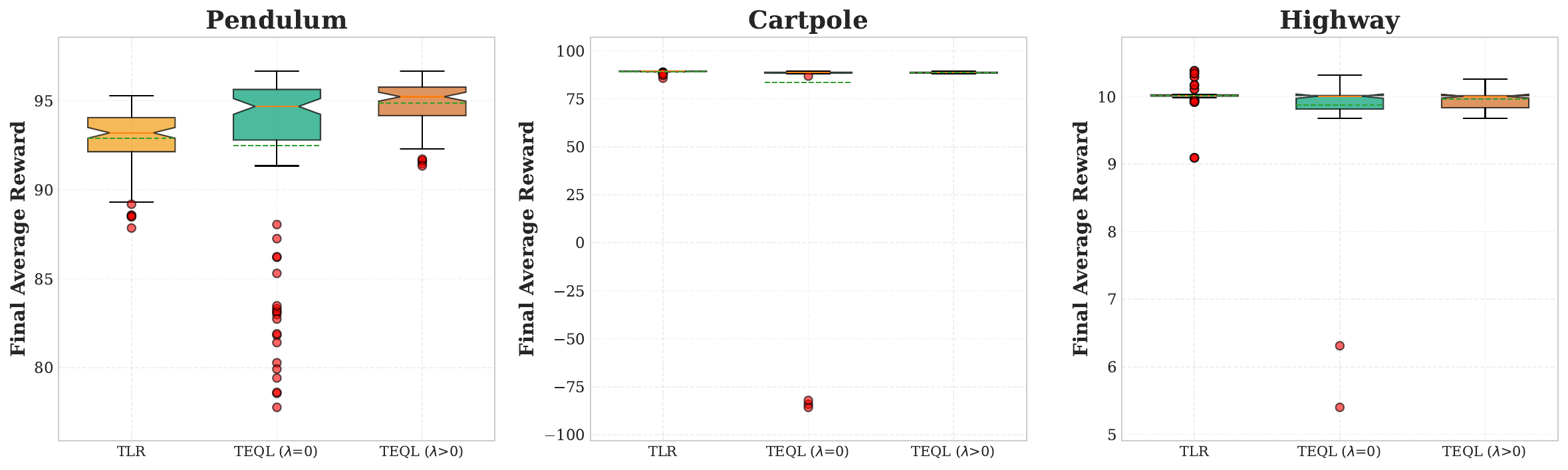}
    \caption{Distribution of final average rewards(last 200 episodes) over 100 runs. Setting $\lambda>0$ substantially reduces standard deviation compared to $\lambda=0$: from 5.24 to 1.30 on Pendulum, from 29.44 to 0.28 on CartPole, and from 0.59 to 0.12 on Highway.}
    \label{fig:ablation_boxplot}
\end{figure}

This section isolates the effect of frequency-aware regularization by comparing TLR, TEQL with $\lambda=0$, and TEQL with $\lambda>0$. All variants share identical CP structure, tensor rank, and TD update rules; only the regularization coefficient differs. Figure~\ref{fig:ablation_curve} shows learning curves, Figure~\ref{fig:ablation_boxplot} reports the distribution of cumulative rewards for last 2000 episodes.

The instability observed when $\lambda=0$ reveals a structural 
vulnerability unique to CP factorization: because parameters in factor matrices 
are shared across the entire state-action fiber, updates in under-sampled regions 
propagate globally. This parameter sharing means that without regularization, 
overestimation at rarely visited pairs can corrupt Q-values throughout the tensor. 
The introduction of $\lambda>0$ provides frequency-aware damping that specifically 
counteracts this effect: at sparsely visited pairs where overestimation risk is highest, 
the regularization coefficient $1/(\mathcal{N}_{t-1}+\epsilon)$ is large, 
resisting rapid value changes. As shown in Figure~\ref{fig:ablation_boxplot}, 
the variance reduction scales with dimensionality: higher-dimensional environments 
exhibit sparser visitation patterns, making frequency-aware stabilization 
increasingly critical.

\subsection{Sensitivity to Discretization Granularity}
\label{subsec:sensitivity}
\begin{figure}[h]
\centering
\includegraphics[width=0.9\textwidth]{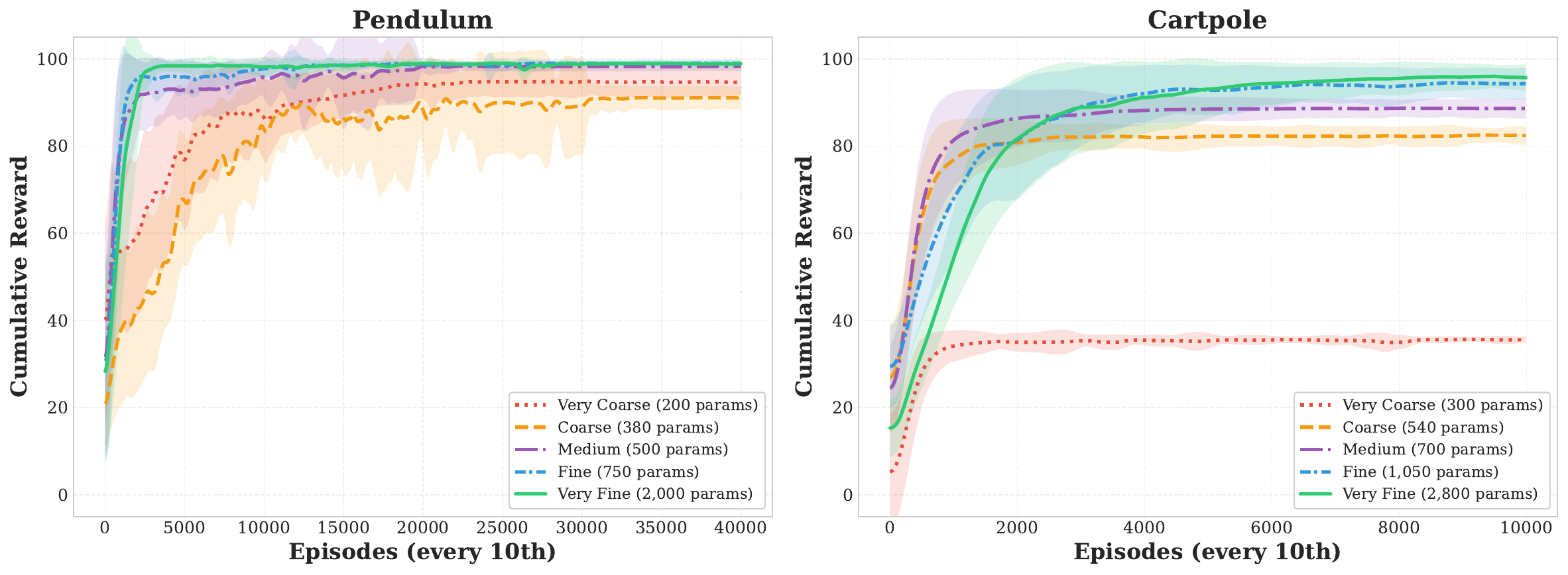} 
\caption{Coarse discretization limits final performance due to approximation bias; fine discretization allows near-optimal convergence. TEQL remains stable across all resolutions. Parameter counts: CartPole 300 to 2,800; Pendulum 200 to 2,000.}
\label{fig:sensitivity}
\end{figure}

\begin{figure}[h]
\centering
\includegraphics[width=0.9\textwidth]{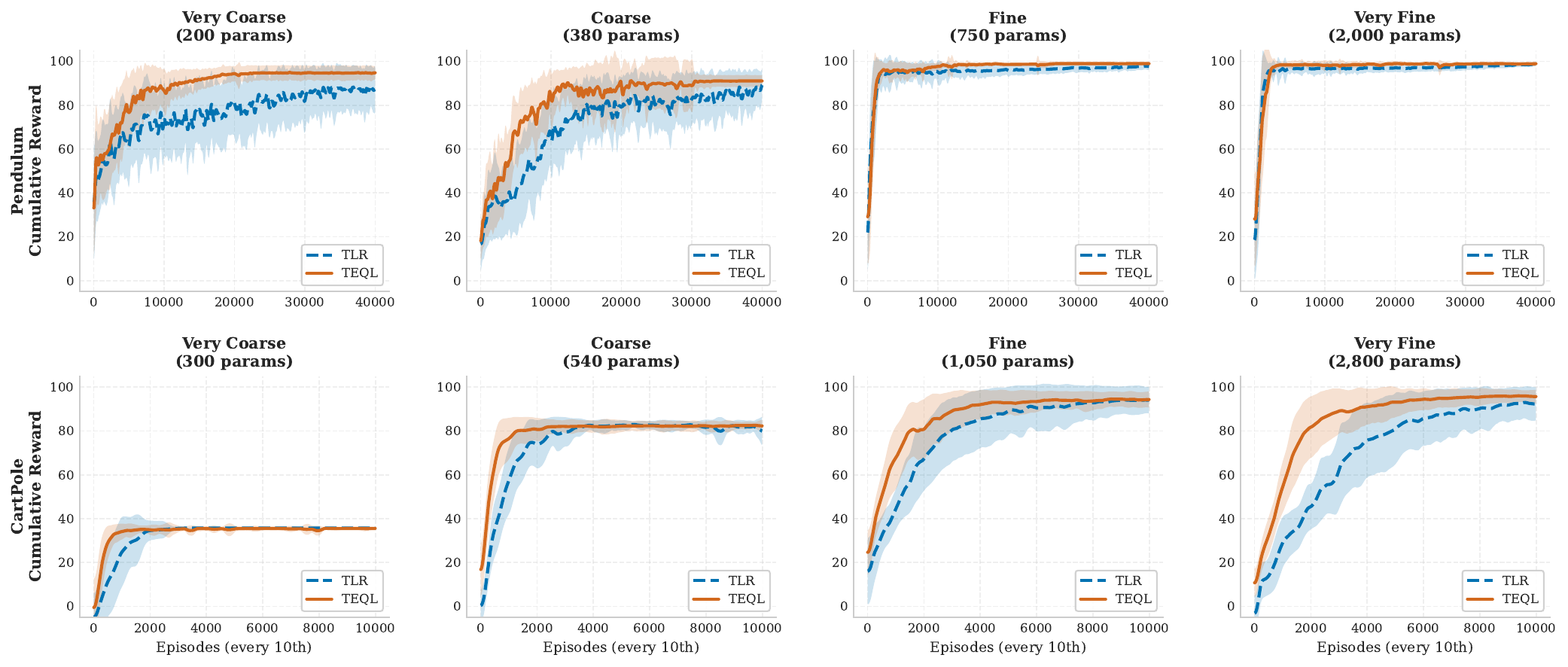} 
\caption{TLR vs TEQL under matched discretization. At coarse 
granularity, both methods are limited by approximation error $B_R$. At fine 
granularity, TEQL exhibits faster convergence and lower variance.}
\label{fig:tlr_teql_sensitivity}
\end{figure}

Performance gains in TEQL are tied to a realizability threshold where discretization resolution matches the requirements of the low-rank model class. Figure~\ref{fig:sensitivity} reveals that under very coarse discretization, TEQL converges reliably but to a suboptimal level. This plateau reflects the approximation bias $B_R$ described in Theorem~\ref{thm:convergence}. Coarse binning collapses distinct dynamics into shared indices and produces a Q-function that lies outside the rank-$R$ model class regardless of sample size. In this regime, the error is dominated by representation bias rather than by stochastic noise.

As resolution increases, performance improves before saturating at near-optimal levels. This transition marks the regime where the structural low-rank assumption becomes approximately satisfied and the convergence guarantees of TEQL become meaningful. Figure~\ref{fig:tlr_teql_sensitivity} compares TEQL and TLR under matched discretization across four granularity levels. At coarse granularity, both methods converge to similar suboptimal solutions because both are limited by $B_R$.
\section{Conclusion}\label{sec-conc}

This work presents TEQL, a framework that uses low-rank tensor structure for sample-efficient reinforcement learning in high-dimensional discrete spaces. The core idea is that low-rank structure in value functions reflects the underlying problem: in many control tasks, state and action variables interact through a limited number of latent factors. This separability serves as a structured inductive bias for learning.

The major contribution of this work lies in reinforcement learning, where TEQL addresses an understudied area: efficient exploration in high-dimensional discrete state-action spaces where data collection is expensive. TEQL leverages tensor decomposition techniques and the EUGE mechanism to improve sample efficiency, which is useful in data-scarce scenarios such as clinical treatment optimization and inventory management. Existing tensor-based Q-learning methods \citep{rozada2024tensor} rely on standard $\varepsilon$-greedy strategies without exploiting structural uncertainty, while continuous methods such as SAC are not designed for problems with a natural discrete structure. EUGE uses tensor reconstruction error as a low-cost measure of estimation uncertainty, and frequency-aware regularization addresses a vulnerability specific to CP-factored value functions: because factor entries are shared along entire state-action fibers, overestimation at a poorly visited pair can propagate globally through the shared parameters. From the tensor analysis perspective, this work provides insights into how tensor decomposition can facilitate reinforcement learning based decision making. The Q-learning problem can be viewed as a tensor estimation problem in a sequential sampling scenario, and the EUGE exploration strategy demonstrates how uncertainty-driven sampling can be leveraged for more efficient tensor completion compared to uniform or random observation patterns.

TEQL is a sample-efficient learning strategy compared to neural network based approaches, and is particularly useful when sampling is expensive or data is scarce in discrete state-action domains. The CP representation achieves a reduction from exponential to linear in the number of dimensions, and provides an interpretable mode-wise factorization. However, since this method is based on the assumption of a low-rank tensor with the rank specified in advance, it is not applicable to problems where the value function does not admit a good low-rank approximation, for example when interactions across dimensions are highly entangled. For naturally continuous cases, as shown in the experiments, the performance improves as the discretization granularity increases, which indicates that for continuous state-action spaces this method might not be the most suitable. The performance of TEQL in such settings is not as strong as methods such as SAC that are designed for continuous cases. However, this does not diminish the contribution of TEQL in discrete action domains where data collection is expensive. The tensor rank $R$ and discretization levels $(d_1, \ldots, d_N)$ are fixed before training; adaptive rank selection or resolution refinement during learning would improve flexibility but is not addressed in this work.

Future research will focus on extending this framework to continuous state and action spaces, where the discrete factor matrices would be replaced by continuous factor functions to eliminate discretization-induced approximation error. This combination of structured decomposition and continuous modeling provides a scalable direction for solving large-scale industrial control and resource optimization problems.

\section{Data availability statement}
The data that support the findings of this study were generated through simulation using standard benchmark environments. The source code implementing the proposed TEQL algorithm, together with scripts for reproducing the experiments and analysis, is openly available in a public repository at https://github.com/Anonymous2025-cmd/teql-anon.

\bibliography{references}

@article{auer2002finite,
  author = {Auer, Peter and Cesa-Bianchi, Nicolò and Fischer, Paul},
  title = {Finite-time analysis of the multiarmed bandit problem},
  journal = {Machine Learning},
  volume = {47},
  number = {2-3},
  pages = {235--256},
  year = {2002}
}

@article{gijsbrechts2022can,
  title={Can deep reinforcement learning improve inventory management? 
          {Performance} on dual sourcing, lost sales, and multi-echelon 
          problems},
   author={Gijsbrechts, Joren and Boute, Robert N and Van Mieghem, 
           Jan A and Zhang, Dennis J},
   journal={Manufacturing \& Service Operations Management},
   volume={24},
   number={3},
   pages={1349--1368},
   year={2022},
   publisher={INFORMS}
 }

@book{wu2011experiments,
   title={Experiments: Planning, Analysis, and Optimization},
   author={Wu, C. F. Jeff and Hamada, Michael S.},
   edition={2},
   year={2011},
   publisher={John Wiley \& Sons}
 }

@article{de2020review,
   title={A review on maintenance optimization},
   author={De Jonge, Bram and Scarf, Philip A},
   journal={European Journal of Operational Research},
   volume={285},
   number={3},
   pages={805--824},
   year={2020},
   publisher={Elsevier}
 }

@inproceedings{thrun1993issues,
  title={Issues in Using Function Approximation for Reinforcement Learning},
  author={Thrun, Sebastian and Schwartz, Anton},
  booktitle={Proceedings of the 1993 Connectionist Models Summer School},
  pages={255--263},
  year={1993},
  publisher={Lawrence Erlbaum Associates},
  address={Hillsdale, NJ}
}

@inproceedings{jiang2017contextual,
  title={Contextual decision processes with low {B}ellman rank are {PAC}-learnable},
  author={Jiang, Nan and Krishnamurthy, Akshay and Agarwal, Alekh and Langford, John and Schapire, Robert E},
  booktitle={Proceedings of the 34th International Conference on Machine Learning},
  series={Proceedings of Machine Learning Research},
  volume={70},
  pages={1704--1713},
  year={2017},
  publisher={PMLR}
}

@inproceedings{agarwal2020flambe,
  title={{FLAMBE}: Structural complexity and representation learning of low rank {MDPs}},
  author={Agarwal, Alekh and Kakade, Sham and Krishnamurthy, Akshay and Sun, Wen},
  booktitle={Advances in Neural Information Processing Systems},
  volume={33},
  pages={20095--20107},
  year={2020},
  publisher={Curran Associates, Inc.}
}

@inproceedings{uehara2022representation,
  title={Representation learning for online and offline {RL} in low-rank {MDPs}},
  author={Uehara, Masatoshi and Zhang, Xuezhou and Sun, Wen},
  booktitle={International Conference on Learning Representations},
  year={2022}
}

@inproceedings{dadashi2019value,
  title={The value function polytope in reinforcement learning},
  author={Dadashi, Robert and Taiga, Adrien Ali and Roux, Nicolas Le and Schuurmans, Dale and Bellemare, Marc G},
  booktitle={International Conference on Machine Learning},
  pages={1486--1495},
  year={2019},
  organization={PMLR}
}

@inproceedings{yang2020harnessing,
  title={Harnessing structures for value-based planning and reinforcement learning},
  author={Yang, Yuzhe and Zhang, Guo and Xu, Zhi-Wei and Katabi, Dina},
  booktitle={International Conference on Learning Representations},
  year={2020}
}

@inproceedings{gheshlaghi2013minimax,
  title={Minimax PAC bounds on the sample complexity of reinforcement learning with a generative model},
  author={Gheshlaghi Azar, Mohammad and Munos, R{\'e}mi and Kappen, Hilbert J},
  booktitle={International Conference on Machine Learning},
  pages={72--80},
  year={2013},
  organization={PMLR}
}

@article{sam2023overcoming,
  title={Overcoming the curse of dimensionality in reinforcement learning through approximate factorization},
  author={Sam, Yohann and Maillard, Odalric-Ambrym and Munos, R{\'e}mi},
  journal={Advances in Neural Information Processing Systems},
  volume={36},
  year={2023}
}

@article{modi2024modelfree,
  title={Model-free representation learning and exploration in low-rank {MDPs}},
  author={Modi, Aditya and Chen, Jinglin and Krishnamurthy, Akshay and Jiang, Nan and Agarwal, Alekh},
  journal={Journal of Machine Learning Research},
  volume={25},
  number={6},
  pages={1--76},
  year={2024}
}

@inproceedings{stojanovic2024lora,
  author    = {Stojanovic, Stefan and Jedra, Yassir and Proutiere, Alexandre},
  title     = {Model-free Low-Rank Reinforcement Learning via Leveraged Entry-wise Matrix Estimation},
  booktitle = {Advances in Neural Information Processing Systems},
  volume    = {37},
  pages     = {30886--30924},
  year      = {2024}
}

@article{christodoulou2019sacdiscrete,
  title   = {Soft Actor-Critic for Discrete Action Settings},
  author  = {Christodoulou, Petros},
  journal = {arXiv preprint arXiv:1910.07207},
  year    = {2019}
}

@inproceedings{stojanovic2024model,
  title={Model-free Low-Rank Reinforcement Learning via Leveraged Entry-wise Matrix Estimation},
  author={Stojanovic, Stefan and Jedra, Yassir and Proutiere, Alexandre},
  booktitle={Advances in Neural Information Processing Systems},
  year={2024}
}

@book{bellman1957dynamic,
  author = {Bellman, Richard},
  title = {Dynamic Programming},
  publisher = {Princeton University Press},
  year = {1957}
}

@book{bertsekas1996neuro,
  author = {Bertsekas, Dimitri P. and Tsitsiklis, John N.},
  title = {Neuro-Dynamic Programming},
  publisher = {Athena Scientific},
  year = {1996}
}

@article{bradtke1996linear,
  author = {Bradtke, Steven J. and Barto, Andrew G.},
  title = {Linear least-squares algorithms for temporal difference learning},
  journal = {Machine Learning},
  volume = {22},
  number = {1-3},
  pages = {33--57},
  year = {1996}
}

@article{jaksch2010near,
  author = {Jaksch, Thomas and Ortner, Ronald and Auer, Peter},
  title = {Near-optimal regret bounds for reinforcement learning},
  journal = {Journal of Machine Learning Research},
  volume = {11},
  pages = {1563--1600},
  year = {2010}
}

@article{mnih2015human,
  author = {Mnih, Volodymyr and Kavukcuoglu, Koray and Silver, David and Rusu, Andrei A. and Veness, Joel and Bellemare, Marc G. and Graves, Alex and Riedmiller, Martin and Fidjeland, Andreas K. and Ostrovski, Georg and others},
  title = {Human-level control through deep reinforcement learning},
  journal = {Nature},
  volume = {518},
  number = {7540},
  pages = {529--533},
  year = {2015}
}

@book{powell2007approximate,
  author = {Powell, Warren B.},
  title = {Approximate Dynamic Programming: Solving the Curses of Dimensionality},
  publisher = {Wiley},
  year = {2007}
}

@book{puterman1994markov,
  author = {Puterman, Martin L.},
  title = {Markov Decision Processes: Discrete Stochastic Dynamic Programming},
  publisher = {Wiley},
  year = {1994}
}

@article{rozada2024tensor,
  author = {Rozada, Santiago and Paternain, Santiago and Marques, Antonio G.},
  title = {Tensor and matrix low-rank value-function approximation in reinforcement learning},
  journal = {IEEE Transactions on Signal Processing},
  volume = {72},
  pages = {1634--1649},
  year = {2024}
}

@misc{shah2020sample,
  author = {Shah, Devavrat and Song, Dogyoon and Xu, Zhi and Yang, Yuzhe},
  title = {Sample efficient reinforcement learning via low-rank matrix estimation},
  howpublished = {arXiv preprint arXiv:2006.01527},
  year = {2020}
}

@book{sutton2018reinforcement,
  author = {Sutton, Richard S. and Barto, Andrew G.},
  title = {Reinforcement Learning: An Introduction},
  edition = {2},
  publisher = {MIT Press},
  year = {2018}
}

@book{szepesvari2010algorithms,
  author = {Szepesvari, Csaba},
  title = {Algorithms for Reinforcement Learning},
  journal = {Synthesis Lectures on Artificial Intelligence and Machine Learning},
  volume = {4},
  number = {1},
  pages = {1--103},
  year = {2010}
}

@article{tsai2021tensor,
  author = {Tsai, Kuan-Chieh and Zhuang, Ziheng and Lent, Ricardo and Wang, Jun and Qi, Qi and Wang, Li-Chun and Han, Zhu},
  title = {Tensor-based reinforcement learning for network routing},
  journal = {IEEE Journal of Selected Topics in Signal Processing},
  volume = {15},
  number = {3},
  pages = {640--653},
  year = {2021}
}

@inproceedings{van2016deep,
  author = {Van Hasselt, Hado and Guez, Arthur and Silver, David},
  title = {Deep reinforcement learning with double Q-learning},
  booktitle = {Proceedings of the AAAI Conference on Artificial Intelligence},
  pages = {2094--2100},
  year = {2016}
}

@article{watkins1992q,
  author = {Watkins, Christopher J. C. H. and Dayan, Peter},
  title = {Q-learning},
  journal = {Machine Learning},
  volume = {8},
  number = {3-4},
  pages = {279--292},
  year = {1992}
}

@article{komorowski2018artificial,
  title={The artificial intelligence clinician learns optimal treatment strategies for sepsis in intensive care},
  author={Komorowski, Matthieu and Celi, Leo A and Badawi, Omar and Gordon, Anthony C and Faisal, A Aldo},
  journal={Nature Medicine},
  volume={24},
  number={11},
  pages={1716--1720},
  year={2018},
  publisher={Nature Publishing Group}
}

@article{liu2019learning,
  title={Learning the dynamic treatment regimes from medical registry data through deep Q-network},
  author={Liu, Ning and Liu, Ying and Logan, Brent and Xu, Zhiyuan and Tang, Jiliang and Wang, Yanzhi},
  journal={Scientific Reports},
  volume={9},
  number={1},
  pages={1--10},
  year={2019},
  publisher={Nature Publishing Group}
}
\if1\anon
\section{Acknowledgment}
This work was supported by the National Science Foundation under Grant No. 2501479.
\fi
\section{Disclosure statement}
The authors have no conflicts of interest to declare.

\phantomsection
\bigskip



\newpage

\appendix
\section{Mathematical Proof}
\label{app:proofs}

This section provides detailed proofs of the theoretical results stated in 
Section~\ref{subsubsec:theoretical_guarantees}. For each result, we first 
restate the claim in precise mathematical form, then provide a complete proof 
with all intermediate steps explicitly shown.

Throughout, we use the notation established in the main text: 
$\hat{\mathcal{Q}}_t$ denotes the TEQL iterate at time $t$, 
$q^*$ is the optimal Q-function, 
$\mathcal{T}$ is the Bellman optimality operator, 
$V_{\max} = R_{\max}/(1-\gamma)$ is the maximal value magnitude, 
$\mathcal{N}_{t-1}(s,a)$ is the visit count of pair $(s,a)$ up to time $t-1$, 
$\Delta\mathcal{Q}_{t-1}(s,a)$ is the decomposition error defined 
in~\eqref{eq:decomp_error}, and $\gamma \in (0,1)$ is the discount factor.

By Assumption~\ref{assump:lowrank}, the optimal Q-function $q^*$ admits a 
rank-$R$ CP approximation with error at most $B_R$. We further assume that 
this approximability extends to Bellman images of bounded low-rank tensors: 
for any rank-$R$ tensor $\mathcal{Q}$ with $\|\mathcal{Q}\|_{\infty} \le V_{\max}$, 
the Bellman image $\mathcal{T}(\mathcal{Q})$ also admits a rank-$R$ approximation 
with error at most $B_R$. This property, known as approximate Bellman closure, 
holds when the MDP transition dynamics exhibit low-rank or approximately 
low-rank structure.


\subsection{Proof of Theorem~\ref{thm:convergence}}
\label{app:convergence_proof}

We first restate Theorem~\ref{thm:convergence} in precise mathematical form.

\textbf{Theorem~\ref{thm:convergence}}[Restatement] Under Assumptions~\ref{assump:lowrank} and~\ref{assump:bounded_iterates}, 
let $\varepsilon_t$ denote the expected magnitude of the stochastic error at 
iteration $t$ arising from using the single-sample TD target~\eqref{eq:target_q} 
instead of the full Bellman expectation. Then the TEQL iterates satisfy, for 
all $t \ge 1$,
\begin{equation}
\mathbb{E}\bigl[\|\hat{\mathcal{Q}}_{t} - q^*\|_{\infty}\bigr]
\le \gamma^t \|\hat{\mathcal{Q}}_{0} - q^*\|_{\infty}
+ \frac{B_R}{1-\gamma}
+ \sum_{k=0}^{t-1} \gamma^{k} \varepsilon_{t-1-k}.
\tag{A.1}
\label{eq:convergence_recursive_app}
\end{equation}
In the limit,
\begin{equation}
\limsup_{t \to \infty} \mathbb{E}\bigl[\|\hat{\mathcal{Q}}_t - q^*\|_{\infty}\bigr]
\le \frac{B_R + \bar{\varepsilon}}{1 - \gamma},
\tag{A.2}
\label{eq:convergence_limit_app}
\end{equation}
where $\bar{\varepsilon} = \limsup_{t \to \infty} \varepsilon_t$.

The bound~\eqref{eq:convergence_recursive_app} decomposes the error into three 
terms: the initial error (decaying geometrically at rate $\gamma$), the 
approximation bias $B_R/(1-\gamma)$ from Assumption~\ref{assump:lowrank}, and 
the accumulated stochastic errors from single-sample TD updates. The asymptotic 
bound~\eqref{eq:convergence_limit_app} shows that TEQL converges to a 
neighborhood of $q^*$ with radius controlled by $B_R$ and the noise level.

\begin{proof}
The proof proceeds in four steps. We first verify that 
Assumption~\ref{assump:bounded_iterates} ensures the iterates remain in a 
domain where the Bellman closure property applies. We then decompose the 
approximation error using the structure of the Bellman operator, establish 
the contraction property, and expand the resulting recursion.

\textbf{Step 1: Iterates Remain in the Bounded Domain.}

By Assumption~\ref{assump:bounded_iterates}, 
$\|\hat{\mathcal{Q}}_t\|_{\infty} \le V_{\max}$ for all $t \ge 0$. This 
assumption can be enforced in practice by projecting or clipping the factor 
matrices $\boldsymbol{F}_n^{(t)}$ after each update.

This boundedness condition ensures that the approximate Bellman closure 
property (stated at the beginning of this appendix) applies at every iteration. 
Specifically, since $\|\hat{\mathcal{Q}}_{t-1}\|_{\infty} \le V_{\max}$, the 
Bellman image $\mathcal{T}(\hat{\mathcal{Q}}_{t-1})$ admits a rank-$R$ 
approximation with error at most $B_R$. This property will be used in Step 3 
to bound the projection error.

\textbf{Step 2: Error Decomposition via Bellman Structure.}

The Bellman optimality operator $\mathcal{T}$ maps any Q-function $\mathcal{Q}$ 
to a new Q-function defined by
\begin{equation}
\mathcal{T}(\mathcal{Q})(s,a) 
= \mathbb{E}\bigl[r + \gamma \max_{a'} \mathcal{Q}(s',a') \,\big|\, s, a\bigr],
\end{equation}
where the expectation is taken over the distribution of rewards $r$ and next 
states $s'$ given the current state-action pair $(s,a)$, as determined by the 
MDP transition kernel $\mathcal{P}$ and reward function $\mathcal{R}$.

A fundamental property of the Bellman optimality operator is that the optimal 
Q-function $q^*$ is its unique fixed point: $\mathcal{T}(q^*) = q^*$. This 
follows from the Bellman optimality equation, which states that for all $(s,a)$,
\begin{equation}
q^*(s,a) = \mathbb{E}\bigl[r + \gamma \max_{a'} q^*(s',a') \,\big|\, s, a\bigr].
\end{equation}

The TEQL update can be modeled abstractly as follows. At each iteration $t$, 
the algorithm performs two operations: (i) it approximates the Bellman image 
$\mathcal{T}(\hat{\mathcal{Q}}_{t-1})$ using a rank-$R$ tensor via block 
coordinate descent, and (ii) it uses a single sampled transition 
$(s_t, a_t, r_t, s_{t+1})$ instead of the full expectation in the Bellman 
operator. We model this abstractly as
\begin{equation}
\hat{\mathcal{Q}}_t = \Pi_R\bigl(\mathcal{T}(\hat{\mathcal{Q}}_{t-1})\bigr) + \zeta_t,
\label{eq:abstract_update_app}
\end{equation}
where $\Pi_R$ denotes a rank-$R$ approximation operator, and $\zeta_t$ captures 
the stochastic error from using the single-sample TD target~\eqref{eq:target_q} 
instead of the full Bellman expectation.

The operator $\Pi_R$ represents the outcome of the block coordinate descent 
procedure in Algorithm~\ref{alg:update}. By the approximate Bellman closure 
property established in Step 1, we have
\begin{equation}
\|\Pi_R(\mathcal{T}(\mathcal{Q})) - \mathcal{T}(\mathcal{Q})\|_{\infty} \le B_R
\label{eq:projection_property}
\end{equation}
for any $\mathcal{Q}$ with $\|\mathcal{Q}\|_{\infty} \le V_{\max}$.

We now decompose the error $\hat{\mathcal{Q}}_t - q^*$. Starting from the 
abstract update~\eqref{eq:abstract_update_app}, we subtract $q^*$ from both sides:
\begin{equation}
\hat{\mathcal{Q}}_t - q^* 
= \Pi_R\bigl(\mathcal{T}(\hat{\mathcal{Q}}_{t-1})\bigr) + \zeta_t - q^*.
\end{equation}

To reveal the structure of this error, we add and subtract intermediate terms. 
First, we add and subtract $\mathcal{T}(\hat{\mathcal{Q}}_{t-1})$:
\begin{align}
\hat{\mathcal{Q}}_t - q^* 
&= \Pi_R\bigl(\mathcal{T}(\hat{\mathcal{Q}}_{t-1})\bigr) 
- \mathcal{T}(\hat{\mathcal{Q}}_{t-1})
+ \mathcal{T}(\hat{\mathcal{Q}}_{t-1})
+ \zeta_t - q^*.
\end{align}

Next, we add and subtract $\mathcal{T}(q^*)$:
\begin{align}
\hat{\mathcal{Q}}_t - q^* 
&= \Pi_R\bigl(\mathcal{T}(\hat{\mathcal{Q}}_{t-1})\bigr) 
- \mathcal{T}(\hat{\mathcal{Q}}_{t-1})
+ \mathcal{T}(\hat{\mathcal{Q}}_{t-1})
- \mathcal{T}(q^*)
+ \mathcal{T}(q^*)
+ \zeta_t - q^*.
\end{align}

Using the fixed point property $\mathcal{T}(q^*) = q^*$, the terms 
$\mathcal{T}(q^*) - q^*$ cancel:
\begin{align}
\hat{\mathcal{Q}}_t - q^* 
&= \underbrace{\Pi_R\bigl(\mathcal{T}(\hat{\mathcal{Q}}_{t-1})\bigr) 
- \mathcal{T}(\hat{\mathcal{Q}}_{t-1})}_{\text{projection error}}
+ \underbrace{\mathcal{T}(\hat{\mathcal{Q}}_{t-1}) - \mathcal{T}(q^*)}_{\text{Bellman contraction term}}
+ \underbrace{\zeta_t}_{\text{stochastic error}}.
\label{eq:error_decomposition}
\end{align}

This decomposition identifies three sources of error: (i)The projection error arises from approximating the Bellman image by a 
rank-$R$ tensor; (ii) The Bellman contraction term propagates the previous iteration's error 
through the Bellman operator. (iii) The stochastic error arises from using a single sampled transition 
instead of the full expectation.

\textbf{Step 3: Contraction Property and One-Step Bound.}

We now establish that the Bellman operator $\mathcal{T}$ is a $\gamma$-contraction 
in the supremum norm. This is a classical result in dynamic programming, but we 
provide the full proof for completeness.

Let $\mathcal{Q}_1$ and $\mathcal{Q}_2$ be any two Q-functions. We want to show that
\begin{equation}
\|\mathcal{T}(\mathcal{Q}_1) - \mathcal{T}(\mathcal{Q}_2)\|_{\infty}
\le \gamma \|\mathcal{Q}_1 - \mathcal{Q}_2\|_{\infty}.
\end{equation}

Fix any state-action pair $(s,a)$. By definition of the Bellman operator,
\begin{align}
\mathcal{T}(\mathcal{Q}_1)(s,a) - \mathcal{T}(\mathcal{Q}_2)(s,a)
&= \mathbb{E}\bigl[r + \gamma \max_{a'} \mathcal{Q}_1(s',a') \,\big|\, s, a\bigr]
- \mathbb{E}\bigl[r + \gamma \max_{a'} \mathcal{Q}_2(s',a') \,\big|\, s, a\bigr].
\end{align}

Since the reward $r$ is determined by the current state-action pair $(s,a)$ 
and does not depend on the Q-function, the $r$ terms cancel:
\begin{align}
\mathcal{T}(\mathcal{Q}_1)(s,a) - \mathcal{T}(\mathcal{Q}_2)(s,a)
&= \gamma \mathbb{E}\bigl[\max_{a'} \mathcal{Q}_1(s',a') \,\big|\, s, a\bigr]
- \gamma \mathbb{E}\bigl[\max_{a'} \mathcal{Q}_2(s',a') \,\big|\, s, a\bigr] \nonumber \\
&= \gamma \mathbb{E}\bigl[\max_{a'} \mathcal{Q}_1(s',a') 
- \max_{a'} \mathcal{Q}_2(s',a') \,\big|\, s, a\bigr],
\end{align}
where we used the linearity of expectation in the last step.

Taking absolute values on both sides:
\begin{equation}
\bigl|\mathcal{T}(\mathcal{Q}_1)(s,a) - \mathcal{T}(\mathcal{Q}_2)(s,a)\bigr|
= \gamma \bigl|\mathbb{E}\bigl[\max_{a'} \mathcal{Q}_1(s',a') 
- \max_{a'} \mathcal{Q}_2(s',a') \,\big|\, s, a\bigr]\bigr|.
\end{equation}

By Jensen's inequality (or equivalently, the triangle inequality for integrals), 
the absolute value of an expectation is bounded by the expectation of the 
absolute value:
\begin{equation}
\bigl|\mathbb{E}[X]\bigr| \le \mathbb{E}\bigl[|X|\bigr]
\end{equation}
for any random variable $X$. Applying this:
\begin{equation}
\bigl|\mathcal{T}(\mathcal{Q}_1)(s,a) - \mathcal{T}(\mathcal{Q}_2)(s,a)\bigr|
\le \gamma \mathbb{E}\bigl[\bigl|\max_{a'} \mathcal{Q}_1(s',a') 
- \max_{a'} \mathcal{Q}_2(s',a')\bigr| \,\big|\, s, a\bigr].
\label{eq:contraction_step1}
\end{equation}

We now bound the term $|\max_{a'} \mathcal{Q}_1(s',a') - \max_{a'} \mathcal{Q}_2(s',a')|$ 
inside the expectation. We claim that for any two vectors 
$x = (x_1, \ldots, x_m)$ and $y = (y_1, \ldots, y_m)$,
\begin{equation}
|\max_i x_i - \max_i y_i| \le \max_i |x_i - y_i|.
\label{eq:max_lipschitz}
\end{equation}

To prove~\eqref{eq:max_lipschitz}, we consider two cases.

\textit{Case 1:} Suppose $\max_i x_i \ge \max_i y_i$. Let $j = \arg\max_i x_i$, 
so $x_j = \max_i x_i$. Then:
\begin{align}
\max_i x_i - \max_i y_i 
&= x_j - \max_i y_i \\
&\le x_j - y_j \quad \text{(since $\max_i y_i \ge y_j$)} \\
&\le |x_j - y_j| \\
&\le \max_i |x_i - y_i|.
\end{align}
Since $\max_i x_i - \max_i y_i \ge 0$ in this case, we have 
$|\max_i x_i - \max_i y_i| = \max_i x_i - \max_i y_i \le \max_i |x_i - y_i|$.

\textit{Case 2:} Suppose $\max_i x_i < \max_i y_i$. By symmetry (swapping the 
roles of $x$ and $y$ in Case 1), we have 
$\max_i y_i - \max_i x_i \le \max_i |y_i - x_i| = \max_i |x_i - y_i|$.
Since $\max_i x_i - \max_i y_i < 0$ in this case, we have 
$|\max_i x_i - \max_i y_i| = \max_i y_i - \max_i x_i \le \max_i |x_i - y_i|$.

In both cases,~\eqref{eq:max_lipschitz} holds.

Applying~\eqref{eq:max_lipschitz} to~\eqref{eq:contraction_step1} with 
$x_{a'} = \mathcal{Q}_1(s', a')$ and $y_{a'} = \mathcal{Q}_2(s', a')$:
\begin{equation}
\bigl|\mathcal{T}(\mathcal{Q}_1)(s,a) - \mathcal{T}(\mathcal{Q}_2)(s,a)\bigr|
\le \gamma \mathbb{E}\bigl[\max_{a'} |\mathcal{Q}_1(s',a') - \mathcal{Q}_2(s',a')| 
\,\big|\, s, a\bigr].
\label{eq:contraction_step2}
\end{equation}

By definition of the supremum norm, for any $s'$:
\begin{equation}
\max_{a'} |\mathcal{Q}_1(s',a') - \mathcal{Q}_2(s',a')| 
\le \sup_{s'',a''} |\mathcal{Q}_1(s'',a'') - \mathcal{Q}_2(s'',a'')|
= \|\mathcal{Q}_1 - \mathcal{Q}_2\|_{\infty}.
\end{equation}

Since $\|\mathcal{Q}_1 - \mathcal{Q}_2\|_{\infty}$ is a constant (independent 
of $s'$), the expectation of a constant equals the constant:
\begin{equation}
\mathbb{E}\bigl[\|\mathcal{Q}_1 - \mathcal{Q}_2\|_{\infty} \,\big|\, s, a\bigr]
= \|\mathcal{Q}_1 - \mathcal{Q}_2\|_{\infty}.
\end{equation}

Substituting into~\eqref{eq:contraction_step2}:
\begin{equation}
\bigl|\mathcal{T}(\mathcal{Q}_1)(s,a) - \mathcal{T}(\mathcal{Q}_2)(s,a)\bigr|
\le \gamma \|\mathcal{Q}_1 - \mathcal{Q}_2\|_{\infty}.
\label{eq:contraction_pointwise}
\end{equation}

The inequality~\eqref{eq:contraction_pointwise} holds for every $(s,a)$. 
Taking the supremum over all $(s,a)$ on the left-hand side:
\begin{equation}
\|\mathcal{T}(\mathcal{Q}_1) - \mathcal{T}(\mathcal{Q}_2)\|_{\infty}
= \sup_{s,a} \bigl|\mathcal{T}(\mathcal{Q}_1)(s,a) - \mathcal{T}(\mathcal{Q}_2)(s,a)\bigr|
\le \gamma \|\mathcal{Q}_1 - \mathcal{Q}_2\|_{\infty}.
\label{eq:contraction_final}
\end{equation}

This establishes that $\mathcal{T}$ is a $\gamma$-contraction in the supremum norm.

We now use the error decomposition~\eqref{eq:error_decomposition} and the 
contraction property~\eqref{eq:contraction_final} to derive a one-step bound.

Taking the supremum norm on both sides of~\eqref{eq:error_decomposition}:
\begin{equation}
\|\hat{\mathcal{Q}}_t - q^*\|_{\infty}
= \bigl\|\bigl[\Pi_R(\mathcal{T}(\hat{\mathcal{Q}}_{t-1})) 
- \mathcal{T}(\hat{\mathcal{Q}}_{t-1})\bigr]
+ \bigl[\mathcal{T}(\hat{\mathcal{Q}}_{t-1}) - \mathcal{T}(q^*)\bigr]
+ \zeta_t\bigr\|_{\infty}.
\end{equation}

Applying the triangle inequality for the supremum norm, which states that 
$\|f + g + h\|_{\infty} \le \|f\|_{\infty} + \|g\|_{\infty} + \|h\|_{\infty}$:
\begin{align}
\|\hat{\mathcal{Q}}_t - q^*\|_{\infty}
&  \le \|\Pi_R(\mathcal{T}(\hat{\mathcal{Q}}_{t-1})) 
- \mathcal{T}(\hat{\mathcal{Q}}_{t-1})\|_{\infty} \nonumber \\
&  + \|\mathcal{T}(\hat{\mathcal{Q}}_{t-1})- \mathcal{T}(q^*)\|_{\infty} \nonumber \\
&  + \|\zeta_t\|_{\infty}.
\label{eq:triangle_inequality_applied}
\end{align}

We bound each of the three terms on the right-hand side.

\textit{Term 1 (projection error):} By Step 1, we have 
$\|\hat{\mathcal{Q}}_{t-1}\|_{\infty} \le V_{\max}$. Therefore, the projection 
property~\eqref{eq:projection_property} applies:
\begin{equation}
\|\Pi_R(\mathcal{T}(\hat{\mathcal{Q}}_{t-1})) 
- \mathcal{T}(\hat{\mathcal{Q}}_{t-1})\|_{\infty} \le B_R.
\label{eq:term1_bound}
\end{equation}

\textit{Term 2 (Bellman contraction):} Applying the contraction 
property~\eqref{eq:contraction_final} with $\mathcal{Q}_1 = \hat{\mathcal{Q}}_{t-1}$ 
and $\mathcal{Q}_2 = q^*$:
\begin{equation}
\|\mathcal{T}(\hat{\mathcal{Q}}_{t-1}) - \mathcal{T}(q^*)\|_{\infty}
\le \gamma \|\hat{\mathcal{Q}}_{t-1} - q^*\|_{\infty}.
\label{eq:term2_bound}
\end{equation}

\textit{Term 3 (stochastic error):} This term is $\|\zeta_t\|_{\infty}$, which we leave as is for now.

Substituting~\eqref{eq:term1_bound} and~\eqref{eq:term2_bound} 
into~\eqref{eq:triangle_inequality_applied}:
\begin{equation}
\|\hat{\mathcal{Q}}_t - q^*\|_{\infty}
\le B_R + \gamma \|\hat{\mathcal{Q}}_{t-1} - q^*\|_{\infty} + \|\zeta_t\|_{\infty}.
\label{eq:one_step_deterministic}
\end{equation}

This is a deterministic inequality that holds for each realization of the 
stochastic process. To obtain a bound on the expected error, we take 
expectations on both sides.

Using the linearity of expectation:
\begin{equation}
\mathbb{E}\bigl[\|\hat{\mathcal{Q}}_t - q^*\|_{\infty}\bigr]
\le \mathbb{E}[B_R] + \gamma \mathbb{E}\bigl[\|\hat{\mathcal{Q}}_{t-1} - q^*\|_{\infty}\bigr] 
+ \mathbb{E}\bigl[\|\zeta_t\|_{\infty}\bigr].
\end{equation}

Since $B_R$ is a constant, $\mathbb{E}[B_R] = B_R$. Defining 
$\varepsilon_t := \mathbb{E}[\|\zeta_t\|_{\infty}]$ as in the theorem statement:
\begin{equation}
\mathbb{E}\bigl[\|\hat{\mathcal{Q}}_t - q^*\|_{\infty}\bigr]
\le B_R + \gamma \mathbb{E}\bigl[\|\hat{\mathcal{Q}}_{t-1} - q^*\|_{\infty}\bigr] 
+ \varepsilon_t.
\label{eq:one_step_expected}
\end{equation}

\textbf{Step 4: Recursive Expansion and Limiting Behavior.}

We now expand the recursion~\eqref{eq:one_step_expected} to obtain explicit 
finite-time and asymptotic bounds.

Define $e_t := \mathbb{E}[\|\hat{\mathcal{Q}}_t - q^*\|_{\infty}]$ for notational 
convenience. The recursion~\eqref{eq:one_step_expected} becomes:
\begin{equation}
e_t \le \gamma e_{t-1} + B_R + \varepsilon_t.
\label{eq:et_recursion}
\end{equation}

We expand this recursion by repeatedly substituting the bound for earlier terms.

\textit{Iteration 1:} Starting from~\eqref{eq:et_recursion}:
\begin{equation}
e_t \le \gamma e_{t-1} + B_R + \varepsilon_t.
\end{equation}

\textit{Iteration 2:} Applying~\eqref{eq:et_recursion} to $e_{t-1}$, we have 
$e_{t-1} \le \gamma e_{t-2} + B_R + \varepsilon_{t-1}$. Substituting:
\begin{align}
e_t &\le \gamma (\gamma e_{t-2} + B_R + \varepsilon_{t-1}) + B_R + \varepsilon_t \\
&= \gamma^2 e_{t-2} + \gamma B_R + B_R + \gamma \varepsilon_{t-1} + \varepsilon_t \\
&= \gamma^2 e_{t-2} + (1 + \gamma) B_R + \gamma \varepsilon_{t-1} + \varepsilon_t.
\end{align}

\textit{Iteration 3:} Applying~\eqref{eq:et_recursion} to $e_{t-2}$, we have 
$e_{t-2} \le \gamma e_{t-3} + B_R + \varepsilon_{t-2}$. Substituting:
\begin{align}
e_t &\le \gamma^2 (\gamma e_{t-3} + B_R + \varepsilon_{t-2}) 
+ (1 + \gamma) B_R + \gamma \varepsilon_{t-1} + \varepsilon_t \\
&= \gamma^3 e_{t-3} + \gamma^2 B_R + (1 + \gamma) B_R 
+ \gamma^2 \varepsilon_{t-2} + \gamma \varepsilon_{t-1} + \varepsilon_t \\
&= \gamma^3 e_{t-3} + (1 + \gamma + \gamma^2) B_R 
+ \gamma^2 \varepsilon_{t-2} + \gamma \varepsilon_{t-1} + \varepsilon_t.
\end{align}

\textit{General pattern:} After $k$ iterations of this expansion, we have:
\begin{equation}
e_t \le \gamma^k e_{t-k} + B_R \sum_{j=0}^{k-1} \gamma^j 
+ \sum_{j=0}^{k-1} \gamma^j \varepsilon_{t-1-j}.
\label{eq:k_step_expansion}
\end{equation}

We verify this pattern by induction. The base case $k=1$ 
is~\eqref{eq:et_recursion}. For the inductive step, assume~\eqref{eq:k_step_expansion} 
holds for some $k$. Applying~\eqref{eq:et_recursion} to $e_{t-k}$:
\begin{align}
e_t &\le \gamma^k (\gamma e_{t-k-1} + B_R + \varepsilon_{t-k}) 
+ B_R \sum_{j=0}^{k-1} \gamma^j + \sum_{j=0}^{k-1} \gamma^j \varepsilon_{t-1-j} \\
&= \gamma^{k+1} e_{t-k-1} + \gamma^k B_R + B_R \sum_{j=0}^{k-1} \gamma^j 
+ \gamma^k \varepsilon_{t-k} + \sum_{j=0}^{k-1} \gamma^j \varepsilon_{t-1-j} \\
&= \gamma^{k+1} e_{t-k-1} + B_R \sum_{j=0}^{k} \gamma^j 
+ \sum_{j=0}^{k} \gamma^j \varepsilon_{t-1-j},
\end{align}
which is~\eqref{eq:k_step_expansion} with $k$ replaced by $k+1$.

\textit{Full expansion:} Setting $k = t$ in~\eqref{eq:k_step_expansion}, so 
that $e_{t-k} = e_0$:
\begin{equation}
e_t \le \gamma^t e_0 + B_R \sum_{j=0}^{t-1} \gamma^j 
+ \sum_{j=0}^{t-1} \gamma^j \varepsilon_{t-1-j}.
\label{eq:full_expansion}
\end{equation}

The initial error $e_0 = \mathbb{E}[\|\hat{\mathcal{Q}}_0 - q^*\|_{\infty}] = 
\|\hat{\mathcal{Q}}_0 - q^*\|_{\infty}$ is deterministic (given the initialization).

For the geometric sum, we use the standard formula. For $\gamma \ne 1$:
\begin{equation}
\sum_{j=0}^{t-1} \gamma^j = \frac{1 - \gamma^t}{1 - \gamma}.
\end{equation}

Since $0 < \gamma < 1$, we have $0 < \gamma^t < 1$, so $1 - \gamma^t < 1$. Therefore:
\begin{equation}
\sum_{j=0}^{t-1} \gamma^j = \frac{1 - \gamma^t}{1 - \gamma} < \frac{1}{1 - \gamma}.
\label{eq:geometric_sum_bound}
\end{equation}

Substituting~\eqref{eq:geometric_sum_bound} into~\eqref{eq:full_expansion}:
\begin{equation}
e_t \le \gamma^t \|\hat{\mathcal{Q}}_0 - q^*\|_{\infty} 
+ \frac{B_R}{1 - \gamma} 
+ \sum_{j=0}^{t-1} \gamma^j \varepsilon_{t-1-j}.
\label{eq:finite_time_bound}
\end{equation}

This establishes the finite-time bound~\eqref{eq:convergence_recursive_app}.

\textit{Asymptotic bound:} We now derive the limiting bound~\eqref{eq:convergence_limit_app} 
by analyzing the behavior of each term in~\eqref{eq:finite_time_bound} as $t \to \infty$.

\textit{Term 1:} Since $0 < \gamma < 1$, we have $\lim_{t \to \infty} \gamma^t = 0$. 
Therefore:
\begin{equation}
\lim_{t \to \infty} \gamma^t \|\hat{\mathcal{Q}}_0 - q^*\|_{\infty} = 0.
\end{equation}

\textit{Term 2:} The term $\frac{B_R}{1-\gamma}$ is a constant, independent of $t$.

\textit{Term 3:} We show that 
$\limsup_{t \to \infty} \sum_{j=0}^{t-1} \gamma^j \varepsilon_{t-1-j} 
\le \frac{\bar{\varepsilon}}{1-\gamma}$, where 
$\bar{\varepsilon} = \limsup_{t \to \infty} \varepsilon_t$.

By the definition of $\limsup$, for any $\eta > 0$, there exists $t_\eta \ge 1$ 
such that $\varepsilon_t \le \bar{\varepsilon} + \eta$ for all $t \ge t_\eta$.

We split the sum $\sum_{j=0}^{t-1} \gamma^j \varepsilon_{t-1-j}$ into two parts 
based on whether the index $t-1-j$ is at least $t_\eta$ or not:
\begin{equation}
\sum_{j=0}^{t-1} \gamma^j \varepsilon_{t-1-j}
= \underbrace{\sum_{j=0}^{t-t_\eta-1} \gamma^j \varepsilon_{t-1-j}}_{\text{Part A}}
+ \underbrace{\sum_{j=t-t_\eta}^{t-1} \gamma^j \varepsilon_{t-1-j}}_{\text{Part B}}.
\label{eq:sum_split}
\end{equation}

\textit{Part A:} For $j$ in the range $0 \le j \le t - t_\eta - 1$, the index 
$t-1-j$ ranges from $(t-1) - 0 = t-1$ down to $(t-1) - (t-t_\eta-1) = t_\eta$. 
Since all these indices are at least $t_\eta$, we have 
$\varepsilon_{t-1-j} \le \bar{\varepsilon} + \eta$ for each term. Therefore:
\begin{align}
\sum_{j=0}^{t-t_\eta-1} \gamma^j \varepsilon_{t-1-j}
&\le \sum_{j=0}^{t-t_\eta-1} \gamma^j (\bar{\varepsilon} + \eta) \\
&= (\bar{\varepsilon} + \eta) \sum_{j=0}^{t-t_\eta-1} \gamma^j \\
&\le (\bar{\varepsilon} + \eta) \sum_{j=0}^{\infty} \gamma^j \\
&= (\bar{\varepsilon} + \eta) \cdot \frac{1}{1 - \gamma} \\
&= \frac{\bar{\varepsilon} + \eta}{1 - \gamma}.
\label{eq:part_a_bound}
\end{align}

\textit{Part B:} For $j$ in the range $t - t_\eta \le j \le t-1$, the index 
$t-1-j$ ranges from $(t-1) - (t-t_\eta) = t_\eta - 1$ down to $(t-1) - (t-1) = 0$. 
Let $M := \max_{0 \le s \le t_\eta - 1} \varepsilon_s$, which is a finite constant 
depending only on $t_\eta$ (and hence only on $\eta$). Then:
\begin{align}
\sum_{j=t-t_\eta}^{t-1} \gamma^j \varepsilon_{t-1-j}
&\le \sum_{j=t-t_\eta}^{t-1} \gamma^j \cdot M \\
&= M \sum_{j=t-t_\eta}^{t-1} \gamma^j.
\end{align}

To evaluate the sum $\sum_{j=t-t_\eta}^{t-1} \gamma^j$, we substitute $k = j - (t - t_\eta)$, 
so $k$ ranges from $0$ to $t_\eta - 1$:
\begin{align}
\sum_{j=t-t_\eta}^{t-1} \gamma^j
&= \sum_{k=0}^{t_\eta-1} \gamma^{k + (t - t_\eta)} \\
&= \gamma^{t - t_\eta} \sum_{k=0}^{t_\eta-1} \gamma^k \\
&\le \gamma^{t - t_\eta} \cdot \frac{1}{1 - \gamma}.
\end{align}

Therefore:
\begin{equation}
\sum_{j=t-t_\eta}^{t-1} \gamma^j \varepsilon_{t-1-j}
\le M \cdot \gamma^{t - t_\eta} \cdot \frac{1}{1 - \gamma}
= \frac{M \gamma^{t - t_\eta}}{1 - \gamma}.
\label{eq:part_b_bound}
\end{equation}

Combining~\eqref{eq:part_a_bound} and~\eqref{eq:part_b_bound}:
\begin{equation}
\sum_{j=0}^{t-1} \gamma^j \varepsilon_{t-1-j}
\le \frac{\bar{\varepsilon} + \eta}{1 - \gamma} + \frac{M \gamma^{t - t_\eta}}{1 - \gamma}.
\label{eq:combined_bound}
\end{equation}

Taking $t \to \infty$ while holding $\eta$ (and hence $t_\eta$ and $M$) fixed: 
since $0 < \gamma < 1$, we have $\gamma^{t - t_\eta} \to 0$ as $t \to \infty$. Therefore:
\begin{equation}
\limsup_{t \to \infty} \sum_{j=0}^{t-1} \gamma^j \varepsilon_{t-1-j}
\le \frac{\bar{\varepsilon} + \eta}{1 - \gamma} + 0
= \frac{\bar{\varepsilon} + \eta}{1 - \gamma}.
\end{equation}

Since this holds for all $\eta > 0$, taking $\eta \to 0$:
\begin{equation}
\limsup_{t \to \infty} \sum_{j=0}^{t-1} \gamma^j \varepsilon_{t-1-j}
\le \frac{\bar{\varepsilon}}{1 - \gamma}.
\label{eq:stochastic_limsup}
\end{equation}

Combining all three terms, we take the $\limsup$ of~\eqref{eq:finite_time_bound}:
\begin{align}
\limsup_{t \to \infty} e_t 
&\le \limsup_{t \to \infty} \gamma^t \|\hat{\mathcal{Q}}_0 - q^*\|_{\infty}
+ \frac{B_R}{1 - \gamma}
+ \limsup_{t \to \infty} \sum_{j=0}^{t-1} \gamma^j \varepsilon_{t-1-j} \\
&\le 0 + \frac{B_R}{1 - \gamma} + \frac{\bar{\varepsilon}}{1 - \gamma} \\
&= \frac{B_R + \bar{\varepsilon}}{1 - \gamma}.
\end{align}

This establishes the asymptotic bound~\eqref{eq:convergence_limit_app} and 
completes the proof of Theorem~\ref{thm:convergence}.
\end{proof}


\subsection{Proof of Proposition~\ref{prop:freq_shrinkage}}
\label{app:freq_shrinkage_proof}

We first restate Proposition~\ref{prop:freq_shrinkage} in precise mathematical form.

\textbf{Proposition~\ref{prop:freq_shrinkage}}[Restatement]
Under Assumption~\ref{assump:bounded_iterates}, suppose the factor matrices 
$\{\boldsymbol{F}_n^{(t)}\}_{n=1}^N$ satisfy $\|\boldsymbol{F}_n^{(t)}\|_{\infty} \le F_{\max}$ 
for all $n$ and $t$ (enforceable via projection). For a single block coordinate 
descent sweep with step size $\alpha_t$, the decomposition error defined 
in~\eqref{eq:decomp_error} satisfies
\begin{equation}
\Delta\mathcal{Q}_t(s_t,a_t)
\le \alpha_t \cdot 2NRF_{\max}^{2(N-1)}V_{\max}
+ \alpha_t \cdot \frac{2\lambda NRF_{\max}^{2(N-1)}V_{\max}}{\mathcal{N}_{t-1}(s_t,a_t) + \epsilon}.
\tag{A.3}
\label{eq:shrinkage_app}
\end{equation}

The bound~\eqref{eq:shrinkage_app} shows that the decomposition error has two 
components: a baseline term proportional to $\alpha_t$ and a term that decreases 
as $1/(\mathcal{N}_{t-1}(s_t,a_t)+\epsilon)$ grows. This formalizes the claim that 
frequency regularization makes updates smaller on frequently visited pairs.

\begin{proof}
We derive the bound by computing the gradient of the TEQL objective with 
respect to the factor matrices and tracking how a single block coordinate 
descent sweep changes the Q-value at the sampled state-action pair. We upper 
bound one full BCD sweep by considering a gradient-type update per mode and 
applying the triangle inequality across modes.

\textbf{Step 1: Setup and Loss Function.}

Fix the sampled pair $(s_t, a_t)$ with corresponding multi-index 
$(i_1, \ldots, i_N)$ in the tensor representation, where $N = D_S + D_A$ is 
the total number of state and action dimensions.

The TEQL per-sample loss function, as given in~\eqref{eq:loss_function}, is
\begin{equation}
\boldsymbol{L}_{s_t,a_t}
= \frac{1}{2}\bigl(q^{\text{target}}_t(s_t,a_t) - \hat{\mathcal{Q}}_t(s_t,a_t)\bigr)^2
- \lambda \frac{\hat{\mathcal{Q}}_t(s_t,a_t)^2}{\mathcal{N}_{t-1}(s_t,a_t) + \epsilon},
\label{eq:loss_restated}
\end{equation}
where:
\begin{itemize}
\item $q^{\text{target}}_t(s_t,a_t) = r_t + \gamma \max_{a'} \hat{\mathcal{Q}}_{t-1}(s_{t+1}, a')$ 
is the TD target defined in~\eqref{eq:target_q}, treated as a constant during 
the factor update;
\item $\lambda > 0$ is the regularization parameter;
\item $\epsilon > 0$ is a small constant to avoid division by zero;
\item $\mathcal{N}_{t-1}(s_t,a_t)$ is the visit count of $(s_t, a_t)$ before time $t$.
\end{itemize}

The first term is the squared TD error, which drives the Q-function toward the 
Bellman target. The second term is the frequency regularizer, which penalizes 
large Q-values at frequently visited state-action pairs by an amount inversely 
proportional to the visit count.

\textbf{Step 2: CP Representation and Gradient Computation.}

By the CP decomposition, the Q-value at $(s_t, a_t)$ is
\begin{equation}
\hat{\mathcal{Q}}_t(s_t,a_t) = \sum_{r=1}^{R} \prod_{n=1}^{N} \boldsymbol{F}_n^{(t)}(i_n, r),
\label{eq:cp_representation}
\end{equation}
where $\boldsymbol{F}_n^{(t)} \in \mathbb{R}^{d_n \times R}$ are the factor matrices 
at time $t$, and $R$ is the CP rank.

Consider updating mode $n$ while holding the other factor matrices 
$\{\boldsymbol{F}_m^{(t)}\}_{m \ne n}$ fixed. We compute the partial derivative of 
$\hat{\mathcal{Q}}_t(s_t,a_t)$ with respect to the factor entry 
$\boldsymbol{F}_n^{(t)}(i_n, r)$.

From~\eqref{eq:cp_representation}, only the $r$-th term in the sum depends on 
$\boldsymbol{F}_n^{(t)}(i_n, r)$:
\begin{equation}
\frac{\partial}{\partial \boldsymbol{F}_n^{(t)}(i_n, r)} 
\sum_{r'=1}^{R} \prod_{m=1}^{N} \boldsymbol{F}_m^{(t)}(i_m, r')
= \frac{\partial}{\partial \boldsymbol{F}_n^{(t)}(i_n, r)} 
\prod_{m=1}^{N} \boldsymbol{F}_m^{(t)}(i_m, r).
\end{equation}

The product $\prod_{m=1}^{N} \boldsymbol{F}_m^{(t)}(i_m, r)$ is linear in 
$\boldsymbol{F}_n^{(t)}(i_n, r)$, so:
\begin{equation}
\frac{\partial}{\partial \boldsymbol{F}_n^{(t)}(i_n, r)} 
\prod_{m=1}^{N} \boldsymbol{F}_m^{(t)}(i_m, r)
= \prod_{m \neq n} \boldsymbol{F}_m^{(t)}(i_m, r).
\label{eq:dQ_dF}
\end{equation}

Now we compute the gradient of the loss~\eqref{eq:loss_restated}. The loss 
consists of two terms, and we differentiate each separately.

\textit{First term (squared TD error):} Let $f(\hat{\mathcal{Q}}) = \frac{1}{2}(q^{\text{target}}_t - \hat{\mathcal{Q}})^2$. 
By the chain rule:
\begin{equation}
\frac{\partial f}{\partial \boldsymbol{F}_n^{(t)}(i_n, r)}
= \frac{\partial f}{\partial \hat{\mathcal{Q}}_t} \cdot 
\frac{\partial \hat{\mathcal{Q}}_t}{\partial \boldsymbol{F}_n^{(t)}(i_n, r)}.
\end{equation}

We have:
\begin{equation}
\frac{\partial f}{\partial \hat{\mathcal{Q}}_t}
= \frac{\partial}{\partial \hat{\mathcal{Q}}_t} 
\left[\frac{1}{2}(q^{\text{target}}_t - \hat{\mathcal{Q}}_t)^2\right]
= (q^{\text{target}}_t - \hat{\mathcal{Q}}_t) \cdot (-1)
= -\bigl(q^{\text{target}}_t - \hat{\mathcal{Q}}_t(s_t,a_t)\bigr).
\end{equation}

Combining with~\eqref{eq:dQ_dF}:
\begin{equation}
\frac{\partial}{\partial \boldsymbol{F}_n^{(t)}(i_n, r)} 
\left[\frac{1}{2}(q^{\text{target}}_t - \hat{\mathcal{Q}}_t)^2\right]
= -\bigl(q^{\text{target}}_t - \hat{\mathcal{Q}}_t(s_t,a_t)\bigr) 
\prod_{m \neq n} \boldsymbol{F}_m^{(t)}(i_m, r).
\label{eq:first_term_gradient}
\end{equation}

\textit{Second term (frequency regularizer):} Let 
$g(\hat{\mathcal{Q}}) = -\lambda \frac{\hat{\mathcal{Q}}^2}{\mathcal{N}_{t-1}(s_t,a_t) + \epsilon}$. 
By the chain rule:
\begin{equation}
\frac{\partial g}{\partial \boldsymbol{F}_n^{(t)}(i_n, r)}
= \frac{\partial g}{\partial \hat{\mathcal{Q}}_t} \cdot 
\frac{\partial \hat{\mathcal{Q}}_t}{\partial \boldsymbol{F}_n^{(t)}(i_n, r)}.
\end{equation}

We have:
\begin{equation}
\frac{\partial g}{\partial \hat{\mathcal{Q}}_t}
= \frac{\partial}{\partial \hat{\mathcal{Q}}_t} 
\left[-\lambda \frac{\hat{\mathcal{Q}}_t^2}{\mathcal{N}_{t-1}(s_t,a_t) + \epsilon}\right]
= -\lambda \cdot \frac{2\hat{\mathcal{Q}}_t}{\mathcal{N}_{t-1}(s_t,a_t) + \epsilon}
= -\frac{2\lambda \hat{\mathcal{Q}}_t(s_t,a_t)}{\mathcal{N}_{t-1}(s_t,a_t) + \epsilon}.
\end{equation}

Combining with~\eqref{eq:dQ_dF}:
\begin{equation}
\frac{\partial}{\partial \boldsymbol{F}_n^{(t)}(i_n, r)} 
\left[-\lambda \frac{\hat{\mathcal{Q}}_t^2}{\mathcal{N}_{t-1}(s_t,a_t) + \epsilon}\right]
= -\frac{2\lambda \hat{\mathcal{Q}}_t(s_t,a_t)}{\mathcal{N}_{t-1}(s_t,a_t) + \epsilon} 
\prod_{m \neq n} \boldsymbol{F}_m^{(t)}(i_m, r).
\label{eq:second_term_gradient}
\end{equation}

\textit{Full gradient:} Adding~\eqref{eq:first_term_gradient} 
and~\eqref{eq:second_term_gradient}:
\begin{align}
\nabla_{\boldsymbol{F}_n(i_n,r)} \boldsymbol{L}_{s_t,a_t}
&= -\bigl(q^{\text{target}}_t - \hat{\mathcal{Q}}_t(s_t,a_t)\bigr) 
\prod_{m \neq n} \boldsymbol{F}_m(i_m, r)
- \frac{2\lambda \hat{\mathcal{Q}}_t(s_t,a_t)}{\mathcal{N}_{t-1}(s_t,a_t) + \epsilon} 
\prod_{m \neq n} \boldsymbol{F}_m(i_m, r) \nonumber \\
&= \left[-\bigl(q^{\text{target}}_t - \hat{\mathcal{Q}}_t(s_t,a_t)\bigr) 
- \frac{2\lambda \hat{\mathcal{Q}}_t(s_t,a_t)}{\mathcal{N}_{t-1}(s_t,a_t) + \epsilon}\right] 
\prod_{m \neq n} \boldsymbol{F}_m(i_m, r) \nonumber \\
&= \left[\bigl(\hat{\mathcal{Q}}_t(s_t,a_t) - q^{\text{target}}_t\bigr) 
- \frac{2\lambda \hat{\mathcal{Q}}_t(s_t,a_t)}{\mathcal{N}_{t-1}(s_t,a_t) + \epsilon}\right] 
\prod_{m \neq n} \boldsymbol{F}_m(i_m, r).
\label{eq:full_gradient}
\end{align}

This matches the gradient formula~\eqref{eq:gradient} in the main text.

\textbf{Step 3: Factor Update and Induced Change in Q-Value.}

By the factor update rule~\eqref{eq:factor_update}, a gradient descent step 
with step size $\alpha_t$ updates the factor entry as:
\begin{equation}
\boldsymbol{F}_n^{(t)}(i_n, r) \leftarrow \boldsymbol{F}_n^{(t)}(i_n, r) 
- \alpha_t \nabla_{\boldsymbol{F}_n(i_n,r)} \boldsymbol{L}_{s_t,a_t}.
\label{eq:factor_update_restated}
\end{equation}

The change in the factor entry is:
\begin{equation}
\Delta \boldsymbol{F}_n^{(t)}(i_n, r) 
:= \boldsymbol{F}_n^{(t),\text{new}}(i_n, r) - \boldsymbol{F}_n^{(t),\text{old}}(i_n, r)
= -\alpha_t \nabla_{\boldsymbol{F}_n(i_n,r)} \boldsymbol{L}_{s_t,a_t}.
\end{equation}

Substituting the gradient~\eqref{eq:full_gradient}:
\begin{equation}
\Delta \boldsymbol{F}_n^{(t)}(i_n, r) 
= -\alpha_t \left[\bigl(\hat{\mathcal{Q}}_t(s_t,a_t) - q^{\text{target}}_t\bigr) 
- \frac{2\lambda \hat{\mathcal{Q}}_t(s_t,a_t)}{\mathcal{N}_{t-1}(s_t,a_t) + \epsilon}\right] 
\prod_{m \neq n} \boldsymbol{F}_m(i_m, r).
\label{eq:delta_F}
\end{equation}

The induced change in $\hat{\mathcal{Q}}_t(s_t,a_t)$ from updating mode $n$ 
(while holding other modes fixed) is computed as follows. By the CP 
representation~\eqref{eq:cp_representation}:
\begin{align}
\Delta_n \hat{\mathcal{Q}}_t(s_t,a_t) 
&:= \hat{\mathcal{Q}}_t^{\text{new}}(s_t,a_t) - \hat{\mathcal{Q}}_t^{\text{old}}(s_t,a_t) \nonumber \\
&= \sum_{r=1}^{R} \left[\prod_{m=1}^{N} \boldsymbol{F}_m^{\text{new}}(i_m, r) 
- \prod_{m=1}^{N} \boldsymbol{F}_m^{\text{old}}(i_m, r)\right].
\end{align}

Since only mode $n$ is updated and all other modes remain fixed, we have 
$\boldsymbol{F}_m^{\text{new}}(i_m, r) = \boldsymbol{F}_m^{\text{old}}(i_m, r)$ 
for $m \ne n$. Therefore:
\begin{align}
\prod_{m=1}^{N} \boldsymbol{F}_m^{\text{new}}(i_m, r) 
- \prod_{m=1}^{N} \boldsymbol{F}_m^{\text{old}}(i_m, r)
&= \boldsymbol{F}_n^{\text{new}}(i_n, r) \prod_{m \neq n} \boldsymbol{F}_m(i_m, r)
- \boldsymbol{F}_n^{\text{old}}(i_n, r) \prod_{m \neq n} \boldsymbol{F}_m(i_m, r) \nonumber \\
&= \bigl[\boldsymbol{F}_n^{\text{new}}(i_n, r) - \boldsymbol{F}_n^{\text{old}}(i_n, r)\bigr] 
\prod_{m \neq n} \boldsymbol{F}_m(i_m, r) \nonumber \\
&= \Delta \boldsymbol{F}_n^{(t)}(i_n, r) \cdot \prod_{m \neq n} \boldsymbol{F}_m(i_m, r).
\end{align}

Summing over $r$:
\begin{equation}
\Delta_n \hat{\mathcal{Q}}_t(s_t,a_t) 
= \sum_{r=1}^{R} \Delta \boldsymbol{F}_n^{(t)}(i_n, r) \cdot \prod_{m \neq n} \boldsymbol{F}_m(i_m, r).
\label{eq:delta_n_Q}
\end{equation}

Substituting~\eqref{eq:delta_F} into~\eqref{eq:delta_n_Q}:
\begin{align}
\Delta_n \hat{\mathcal{Q}}_t(s_t,a_t) 
&= \sum_{r=1}^{R} \left\{-\alpha_t \left[\bigl(\hat{\mathcal{Q}}_t - q^{\text{target}}_t\bigr) 
- \frac{2\lambda \hat{\mathcal{Q}}_t}{\mathcal{N}_{t-1} + \epsilon}\right] 
\prod_{m \neq n} \boldsymbol{F}_m(i_m, r)\right\} \cdot \prod_{m \neq n} \boldsymbol{F}_m(i_m, r) \nonumber \\
&= -\alpha_t \left[\bigl(\hat{\mathcal{Q}}_t - q^{\text{target}}_t\bigr) 
- \frac{2\lambda \hat{\mathcal{Q}}_t}{\mathcal{N}_{t-1} + \epsilon}\right] 
\sum_{r=1}^{R} \left(\prod_{m \neq n} \boldsymbol{F}_m(i_m, r)\right)^2,
\label{eq:delta_n_expanded}
\end{align}
where we abbreviated $\hat{\mathcal{Q}}_t = \hat{\mathcal{Q}}_t(s_t,a_t)$ and 
$\mathcal{N}_{t-1} = \mathcal{N}_{t-1}(s_t,a_t)$ for readability.

Taking absolute values:
\begin{equation}
|\Delta_n \hat{\mathcal{Q}}_t(s_t,a_t)| 
= \alpha_t \left|\bigl(\hat{\mathcal{Q}}_t - q^{\text{target}}_t\bigr) 
- \frac{2\lambda \hat{\mathcal{Q}}_t}{\mathcal{N}_{t-1} + \epsilon}\right| 
\sum_{r=1}^{R} \left(\prod_{m \neq n} \boldsymbol{F}_m(i_m, r)\right)^2.
\label{eq:delta_n_abs}
\end{equation}

\textbf{Step 4: Bounding the Product of Factor Entries.}

We now bound $\sum_{r=1}^{R} (\prod_{m \neq n} \boldsymbol{F}_m(i_m, r))^2$.

By assumption, each factor entry satisfies 
$|\boldsymbol{F}_m^{(t)}(i_m, r)| \le F_{\max}$ for all $m$, $r$, and $t$. 
Therefore, for each $r$:
\begin{equation}
\left|\prod_{m \neq n} \boldsymbol{F}_m(i_m, r)\right|
= \prod_{m \neq n} |\boldsymbol{F}_m(i_m, r)|
\le \prod_{m \neq n} F_{\max}
= F_{\max}^{N-1},
\end{equation}
where the product is over $N-1$ terms (all modes except mode $n$).

Squaring both sides:
\begin{equation}
\left(\prod_{m \neq n} \boldsymbol{F}_m(i_m, r)\right)^2 \le F_{\max}^{2(N-1)}.
\end{equation}

Summing over $r = 1, \ldots, R$:
\begin{equation}
\sum_{r=1}^{R} \left(\prod_{m \neq n} \boldsymbol{F}_m(i_m, r)\right)^2 
\le \sum_{r=1}^{R} F_{\max}^{2(N-1)}
= R \cdot F_{\max}^{2(N-1)}.
\label{eq:sum_bound}
\end{equation}

\textbf{Step 5: Bounding the Bracket Term.}

We now bound the term 
$\left|\bigl(\hat{\mathcal{Q}}_t - q^{\text{target}}_t\bigr) 
- \frac{2\lambda \hat{\mathcal{Q}}_t}{\mathcal{N}_{t-1} + \epsilon}\right|$ 
in~\eqref{eq:delta_n_abs}.

By the triangle inequality, for any real numbers $a$ and $b$:
\begin{equation}
|a - b| \le |a| + |b|.
\end{equation}

Applying this with $a = \hat{\mathcal{Q}}_t - q^{\text{target}}_t$ and 
$b = \frac{2\lambda \hat{\mathcal{Q}}_t}{\mathcal{N}_{t-1} + \epsilon}$:
\begin{equation}
\left|\bigl(\hat{\mathcal{Q}}_t - q^{\text{target}}_t\bigr) 
- \frac{2\lambda \hat{\mathcal{Q}}_t}{\mathcal{N}_{t-1} + \epsilon}\right|
\le |\hat{\mathcal{Q}}_t - q^{\text{target}}_t| 
+ \frac{2\lambda |\hat{\mathcal{Q}}_t|}{\mathcal{N}_{t-1} + \epsilon}.
\label{eq:bracket_triangle}
\end{equation}

We bound each term on the right-hand side separately.

\textit{Bounding $|\hat{\mathcal{Q}}_t(s_t,a_t)|$:} By 
Assumption~\ref{assump:bounded_iterates}, 
$\|\hat{\mathcal{Q}}_t\|_{\infty} \le V_{\max}$, so:
\begin{equation}
|\hat{\mathcal{Q}}_t(s_t,a_t)| \le V_{\max}.
\label{eq:Q_bound}
\end{equation}

\textit{Bounding $|q^{\text{target}}_t(s_t,a_t)|$:} By definition~\eqref{eq:target_q}:
\begin{equation}
q^{\text{target}}_t(s_t,a_t) = r_t + \gamma \max_{a'} \hat{\mathcal{Q}}_{t-1}(s_{t+1}, a').
\end{equation}

Taking absolute values and applying the triangle inequality:
\begin{align}
|q^{\text{target}}_t(s_t,a_t)| 
&= |r_t + \gamma \max_{a'} \hat{\mathcal{Q}}_{t-1}(s_{t+1}, a')| \nonumber \\
&\le |r_t| + \gamma |\max_{a'} \hat{\mathcal{Q}}_{t-1}(s_{t+1}, a')|.
\end{align}

By the bounded rewards assumption, $|r_t| \le R_{\max}$. By 
Assumption~\ref{assump:bounded_iterates}, 
$|\max_{a'} \hat{\mathcal{Q}}_{t-1}(s_{t+1}, a')| \le \|\hat{\mathcal{Q}}_{t-1}\|_{\infty} \le V_{\max}$. 
Therefore:
\begin{equation}
|q^{\text{target}}_t(s_t,a_t)| \le R_{\max} + \gamma V_{\max}.
\label{eq:target_bound}
\end{equation}

\textit{Bounding $|\hat{\mathcal{Q}}_t - q^{\text{target}}_t|$:} Applying the 
triangle inequality:
\begin{align}
|\hat{\mathcal{Q}}_t(s_t,a_t) - q^{\text{target}}_t(s_t,a_t)| 
&\le |\hat{\mathcal{Q}}_t(s_t,a_t)| + |q^{\text{target}}_t(s_t,a_t)| \nonumber \\
&\le V_{\max} + (R_{\max} + \gamma V_{\max}) \nonumber \\
&= V_{\max} + R_{\max} + \gamma V_{\max}.
\label{eq:td_error_intermediate}
\end{align}

We simplify using the relationship $V_{\max} = R_{\max}/(1-\gamma)$, which 
implies $R_{\max} = (1-\gamma) V_{\max}$. Substituting:
\begin{align}
V_{\max} + R_{\max} + \gamma V_{\max}
&= V_{\max} + (1-\gamma) V_{\max} + \gamma V_{\max} \nonumber \\
&= V_{\max} + V_{\max} - \gamma V_{\max} + \gamma V_{\max} \nonumber \\
&= 2 V_{\max}.
\label{eq:td_error_simplified}
\end{align}

Therefore:
\begin{equation}
|\hat{\mathcal{Q}}_t(s_t,a_t) - q^{\text{target}}_t(s_t,a_t)| \le 2 V_{\max}.
\label{eq:td_error_final}
\end{equation}

Substituting~\eqref{eq:Q_bound} and~\eqref{eq:td_error_final} 
into~\eqref{eq:bracket_triangle}:
\begin{equation}
\left|\bigl(\hat{\mathcal{Q}}_t - q^{\text{target}}_t\bigr) 
- \frac{2\lambda \hat{\mathcal{Q}}_t}{\mathcal{N}_{t-1} + \epsilon}\right|
\le 2 V_{\max} + \frac{2\lambda V_{\max}}{\mathcal{N}_{t-1}(s_t,a_t) + \epsilon}.
\label{eq:bracket_final}
\end{equation}

\textbf{Step 6: Bounding the Change from Mode $n$.}

Substituting~\eqref{eq:sum_bound} and~\eqref{eq:bracket_final} 
into~\eqref{eq:delta_n_abs}:
\begin{align}
|\Delta_n \hat{\mathcal{Q}}_t(s_t,a_t)| 
&\le \alpha_t \left(2 V_{\max} + \frac{2\lambda V_{\max}}{\mathcal{N}_{t-1}(s_t,a_t) + \epsilon}\right) 
\cdot R \cdot F_{\max}^{2(N-1)} \nonumber \\
&= \alpha_t \cdot R F_{\max}^{2(N-1)} \cdot 2 V_{\max} 
+ \alpha_t \cdot R F_{\max}^{2(N-1)} \cdot \frac{2\lambda V_{\max}}{\mathcal{N}_{t-1}(s_t,a_t) + \epsilon}.
\label{eq:delta_n_final}
\end{align}

\textbf{Step 7: Summing Over All Modes.}

A full block coordinate descent sweep updates all $N$ modes sequentially. 
Although the factor entries change as earlier modes are updated, the bound 
$|\boldsymbol{F}_m(i_m, r)| \le F_{\max}$ holds uniformly throughout the sweep 
(by the boundedness assumption or by enforcing projection after each mode update). 
Therefore, the bound~\eqref{eq:delta_n_final} applies to each mode update.

The total change in $\hat{\mathcal{Q}}_t(s_t,a_t)$ from the full sweep is:
\begin{equation}
\Delta\mathcal{Q}_t(s_t,a_t) 
= |\hat{\mathcal{Q}}_{t}(s_t,a_t) - \hat{\mathcal{Q}}_{t-1}(s_t,a_t)|,
\end{equation}
as defined in~\eqref{eq:decomp_error}.

The change can be written as the sum of changes from each mode. By the triangle 
inequality:
\begin{equation}
\Delta\mathcal{Q}_t(s_t,a_t) 
= \left|\sum_{n=1}^{N} \Delta_n \hat{\mathcal{Q}}_t(s_t,a_t)\right|
\le \sum_{n=1}^{N} |\Delta_n \hat{\mathcal{Q}}_t(s_t,a_t)|.
\label{eq:total_triangle}
\end{equation}

Substituting~\eqref{eq:delta_n_final} for each mode:
\begin{align}
\Delta\mathcal{Q}_t(s_t,a_t) 
&\le \sum_{n=1}^{N} \left[\alpha_t \cdot R F_{\max}^{2(N-1)} \cdot 2 V_{\max} 
+ \alpha_t \cdot R F_{\max}^{2(N-1)} \cdot \frac{2\lambda V_{\max}}{\mathcal{N}_{t-1}(s_t,a_t) + \epsilon}\right] \nonumber \\
&= N \cdot \alpha_t \cdot R F_{\max}^{2(N-1)} \cdot 2 V_{\max} 
+ N \cdot \alpha_t \cdot R F_{\max}^{2(N-1)} \cdot \frac{2\lambda V_{\max}}{\mathcal{N}_{t-1}(s_t,a_t) + \epsilon} \nonumber \\
&= \alpha_t \cdot 2 N R F_{\max}^{2(N-1)} V_{\max} 
+ \alpha_t \cdot \frac{2\lambda N R F_{\max}^{2(N-1)} V_{\max}}{\mathcal{N}_{t-1}(s_t,a_t) + \epsilon}.
\label{eq:total_final}
\end{align}

This is exactly the bound~\eqref{eq:shrinkage_app}, completing the proof.
\end{proof}


\subsection{Proof of Proposition~\ref{prop:euge_selection}}
\label{app:euge_selection_proof}

We first restate Proposition~\ref{prop:euge_selection} in precise mathematical form.

\textbf{Proposition~\ref{prop:euge_selection}}[Restatement]
Fix a state $s \in \mathcal{S}$ and an action $a \in \mathcal{A}$. Recall the 
EUGE selection rule~\eqref{eq:euge}:
\begin{equation}
\mathrm{EU}_t(s,a)
= \hat{\mathcal{Q}}_{t-1}(s,a)
+ c\left(\Delta\mathcal{Q}_{t-1}(s,a)
+ \sqrt{\frac{\log \mathcal{N}_{\mathrm{total},t-1}(s)}{\mathcal{N}_{t-1}(s,a) + 1}}\right),
\end{equation}
where $\mathcal{N}_{\mathrm{total},t-1}(s) = \sum_{a'} \mathcal{N}_{t-1}(s,a')$ 
is the total visits to state $s$, and $c > 0$ is the exploration parameter. 
Suppose there exist $\delta > 0$ and $t_0 \ge 1$ such that for all $t \ge t_0$:
\begin{enumerate}
\item[(i)] $\hat{\mathcal{Q}}_{t-1}(s,a) \le \max_{a'} \hat{\mathcal{Q}}_{t-1}(s,a') - \delta$ 
\quad (suboptimality margin);
\item[(ii)] $c \cdot \Delta\mathcal{Q}_{t-1}(s,a) \le \delta/4$ 
\quad (small decomposition-error bonus).
\end{enumerate}
Then the total number of times $a$ is selected at $s$ up to horizon $T$ satisfies
\begin{equation}
\sum_{t=1}^{T} \mathbf{1}\{s_t = s,\, a_t = a\} \le t_0 + \frac{16 c^2}{\delta^2} \log T.
\tag{A.4}
\label{eq:selection_app}
\end{equation}

The bound~\eqref{eq:selection_app} shows that actions satisfying conditions 
(i)-(ii) are selected only $O(\log T)$ times. This formalizes the claim that 
EUGE avoids wasteful re-selection of suboptimal low-uncertainty actions.

\begin{proof}
The proof uses a counting argument based on the EUGE index structure 
defined in~\eqref{eq:euge}. The key insight is that the UCB-type exploration 
bonus decays as an action accumulates visits, which limits how often such an 
action can be chosen.

\textbf{Step 1: EUGE Selection Rule.}

By the EUGE rule in Algorithm~\ref{alg:euge}, the action selected at time $t$ 
in state $s_t$ is:
\begin{equation}
a_t = \arg\max_{a' \in \mathcal{A}} \mathrm{EU}_t(s_t, a'),
\end{equation}
where the EUGE score is defined as:
\begin{equation}
\mathrm{EU}_t(s_t, a')
= \hat{\mathcal{Q}}_{t-1}(s_t, a')
+ c\left(\Delta\mathcal{Q}_{t-1}(s_t, a')
+ \sqrt{\frac{\log \mathcal{N}_{\mathrm{total},t-1}(s_t)}{\mathcal{N}_{t-1}(s_t, a') + 1}}\right).
\end{equation}

The EUGE score consists of three components:
\begin{itemize}
\item The estimated Q-value $\hat{\mathcal{Q}}_{t-1}(s_t, a')$;
\item The decomposition error bonus $c \cdot \Delta\mathcal{Q}_{t-1}(s_t, a')$, 
which captures uncertainty from recent changes in the Q-estimate;
\item The UCB-type bonus $c \sqrt{\frac{\log \mathcal{N}_{\mathrm{total},t-1}(s_t)}{\mathcal{N}_{t-1}(s_t, a') + 1}}$, 
which encourages exploration of less-visited actions.
\end{itemize}

\textbf{Step 2: Selection Implies Large EUGE Bonus.}

Suppose action $a$ is selected at state $s$ at time $t$. Let 
$a^* = \arg\max_{a'} \hat{\mathcal{Q}}_{t-1}(s, a')$ denote the greedy action 
(the action with the highest estimated Q-value).

Since $a$ is selected, its EUGE score must be at least as large as that of 
any other action, including $a^*$:
\begin{equation}
\mathrm{EU}_t(s, a) \ge \mathrm{EU}_t(s, a^*).
\label{eq:euge_inequality}
\end{equation}

We derive a lower bound on $\mathrm{EU}_t(s, a^*)$. By definition:
\begin{equation}
\mathrm{EU}_t(s, a^*)
= \hat{\mathcal{Q}}_{t-1}(s, a^*)
+ c\left(\Delta\mathcal{Q}_{t-1}(s, a^*)
+ \sqrt{\frac{\log \mathcal{N}_{\mathrm{total},t-1}(s)}{\mathcal{N}_{t-1}(s, a^*) + 1}}\right).
\end{equation}

Since both $\Delta\mathcal{Q}_{t-1}(s, a^*) \ge 0$ (by definition as an absolute 
value) and $\sqrt{\frac{\log \mathcal{N}_{\mathrm{total},t-1}(s)}{\mathcal{N}_{t-1}(s, a^*) + 1}} \ge 0$, 
the EUGE bonus is non-negative:
\begin{equation}
c\left(\Delta\mathcal{Q}_{t-1}(s, a^*)
+ \sqrt{\frac{\log \mathcal{N}_{\mathrm{total},t-1}(s)}{\mathcal{N}_{t-1}(s, a^*) + 1}}\right) \ge 0.
\end{equation}

Therefore:
\begin{equation}
\mathrm{EU}_t(s, a^*) \ge \hat{\mathcal{Q}}_{t-1}(s, a^*).
\label{eq:eu_star_lower}
\end{equation}

Combining~\eqref{eq:euge_inequality} and~\eqref{eq:eu_star_lower}:
\begin{equation}
\mathrm{EU}_t(s, a) \ge \hat{\mathcal{Q}}_{t-1}(s, a^*).
\label{eq:eu_a_lower}
\end{equation}

Expanding $\mathrm{EU}_t(s, a)$:
\begin{equation}
\hat{\mathcal{Q}}_{t-1}(s, a)
+ c\left(\Delta\mathcal{Q}_{t-1}(s, a)
+ \sqrt{\frac{\log \mathcal{N}_{\mathrm{total},t-1}(s)}{\mathcal{N}_{t-1}(s, a) + 1}}\right)
\ge \hat{\mathcal{Q}}_{t-1}(s, a^*).
\end{equation}

Rearranging to isolate the EUGE bonus:
\begin{equation}
c\left(\Delta\mathcal{Q}_{t-1}(s, a)
+ \sqrt{\frac{\log \mathcal{N}_{\mathrm{total},t-1}(s)}{\mathcal{N}_{t-1}(s, a) + 1}}\right)
\ge \hat{\mathcal{Q}}_{t-1}(s, a^*) - \hat{\mathcal{Q}}_{t-1}(s, a).
\label{eq:bonus_lower}
\end{equation}

\textbf{Step 3: Applying Conditions (i) and (ii).}

Now suppose $t \ge t_0$ and both conditions (i) and (ii) hold.

By condition (i), the value gap satisfies:
\begin{equation}
\hat{\mathcal{Q}}_{t-1}(s, a) \le \max_{a'} \hat{\mathcal{Q}}_{t-1}(s, a') - \delta
= \hat{\mathcal{Q}}_{t-1}(s, a^*) - \delta.
\end{equation}

Rearranging:
\begin{equation}
\hat{\mathcal{Q}}_{t-1}(s, a^*) - \hat{\mathcal{Q}}_{t-1}(s, a) \ge \delta.
\label{eq:value_gap}
\end{equation}

Substituting~\eqref{eq:value_gap} into~\eqref{eq:bonus_lower}: if action $a$ 
is selected at time $t \ge t_0$, then:
\begin{equation}
c\left(\Delta\mathcal{Q}_{t-1}(s, a)
+ \sqrt{\frac{\log \mathcal{N}_{\mathrm{total},t-1}(s)}{\mathcal{N}_{t-1}(s, a) + 1}}\right)
\ge \delta.
\label{eq:total_bonus_delta}
\end{equation}

By condition (ii), the decomposition-error bonus is bounded:
\begin{equation}
c \cdot \Delta\mathcal{Q}_{t-1}(s, a) \le \frac{\delta}{4}.
\label{eq:delta_Q_bounded}
\end{equation}

Substituting~\eqref{eq:delta_Q_bounded} into~\eqref{eq:total_bonus_delta}:
\begin{align}
\delta  \,& \le c\left(\Delta\mathcal{Q}_{t-1}(s, a)
+ \sqrt{\frac{\log \mathcal{N}_{\mathrm{total},t-1}(s)}{\mathcal{N}_{t-1}(s, a) + 1}}\right) \nonumber \\
&= c \cdot \Delta\mathcal{Q}_{t-1}(s, a)
+ c \sqrt{\frac{\log \mathcal{N}_{\mathrm{total},t-1}(s)}{\mathcal{N}_{t-1}(s, a) + 1}} \nonumber \\
&\le \frac{\delta}{4}
+ c \sqrt{\frac{\log \mathcal{N}_{\mathrm{total},t-1}(s)}{\mathcal{N}_{t-1}(s, a) + 1}}.
\end{align}

Rearranging to isolate the UCB term:
\begin{equation}
c \sqrt{\frac{\log \mathcal{N}_{\mathrm{total},t-1}(s)}{\mathcal{N}_{t-1}(s, a) + 1}}
\ge \delta - \frac{\delta}{4}
= \frac{4\delta - \delta}{4}
= \frac{3\delta}{4}.
\label{eq:ucb_lower}
\end{equation}

\textbf{Step 4: Converting to a Visit Count Bound.}

We now convert the inequality~\eqref{eq:ucb_lower} into an upper bound on 
the visit count $\mathcal{N}_{t-1}(s, a)$.

Squaring both sides of~\eqref{eq:ucb_lower} (both sides are non-negative):
\begin{equation}
c^2 \cdot \frac{\log \mathcal{N}_{\mathrm{total},t-1}(s)}{\mathcal{N}_{t-1}(s, a) + 1}
\ge \frac{9\delta^2}{16}.
\label{eq:squared_inequality}
\end{equation}

Rearranging to solve for $\mathcal{N}_{t-1}(s, a) + 1$: multiply both sides 
by $(\mathcal{N}_{t-1}(s, a) + 1)$ and divide by $\frac{9\delta^2}{16}$:
\begin{equation}
\mathcal{N}_{t-1}(s, a) + 1 
\le \frac{c^2 \log \mathcal{N}_{\mathrm{total},t-1}(s)}{\frac{9\delta^2}{16}}
= \frac{16 c^2 \log \mathcal{N}_{\mathrm{total},t-1}(s)}{9 \delta^2}.
\end{equation}

Subtracting 1 from both sides:
\begin{equation}
\mathcal{N}_{t-1}(s, a) 
\le \frac{16 c^2 \log \mathcal{N}_{\mathrm{total},t-1}(s)}{9 \delta^2} - 1
< \frac{16 c^2 \log \mathcal{N}_{\mathrm{total},t-1}(s)}{9 \delta^2}.
\label{eq:N_upper_t}
\end{equation}

Since the total visits to state $s$ cannot exceed the total number of time 
steps, we have $\mathcal{N}_{\mathrm{total},t-1}(s) \le t - 1 \le T$ for $t \le T$. 
Taking logarithms (which is monotone increasing):
\begin{equation}
\log \mathcal{N}_{\mathrm{total},t-1}(s) \le \log T.
\end{equation}

Substituting into~\eqref{eq:N_upper_t}: selection at any $t \in [t_0, T]$ requires:
\begin{equation}
\mathcal{N}_{t-1}(s, a) < \frac{16 c^2 \log T}{9 \delta^2}.
\label{eq:N_bound}
\end{equation}

\textbf{Step 5: Counting Total Selections.}

Let $n_T(s, a) := \sum_{t=1}^{T} \mathbf{1}\{s_t = s,\, a_t = a\}$ denote the 
total number of times action $a$ is selected at state $s$ up to time $T$.

We split this count into two periods: before time $t_0$ and from $t_0$ to $T$:
\begin{equation}
n_T(s, a) = n_{t_0-1}(s, a) + \bigl[n_T(s, a) - n_{t_0-1}(s, a)\bigr],
\label{eq:count_split}
\end{equation}
where $n_{t_0-1}(s, a) = \sum_{t=1}^{t_0-1} \mathbf{1}\{s_t = s,\, a_t = a\}$ 
is the number of selections before time $t_0$.

\textit{Bound on selections before $t_0$:} Since at most one action can be 
selected per time step, and there are $t_0 - 1$ time steps before $t_0$:
\begin{equation}
n_{t_0-1}(s, a) \le t_0 - 1.
\label{eq:before_t0}
\end{equation}

\textit{Bound on selections from $t_0$ to $T$:} For $t \in [t_0, T]$, we use 
the visit count bound~\eqref{eq:N_bound}.

Each time action $a$ is selected at state $s$, the visit count 
$\mathcal{N}_t(s, a)$ increases by 1. Selection at time $t$ can only occur if 
$\mathcal{N}_{t-1}(s, a) < \frac{16 c^2 \log T}{9 \delta^2}$.

Starting from $\mathcal{N}_{t_0-1}(s, a) \ge 0$, consider the sequence of 
selections at times $t_0 \le t_{(1)} < t_{(2)} < \cdots \le T$. After the 
$k$-th selection (at time $t_{(k)}$), we have 
$\mathcal{N}_{t_{(k)}}(s, a) = \mathcal{N}_{t_{(k)}-1}(s, a) + 1 \ge k$.

For the $(k+1)$-th selection to occur at some time $t_{(k+1)} > t_{(k)}$, we 
need $\mathcal{N}_{t_{(k+1)}-1}(s, a) < \frac{16 c^2 \log T}{9 \delta^2}$. 
Since $\mathcal{N}_{t_{(k+1)}-1}(s, a) \ge \mathcal{N}_{t_{(k)}}(s, a) \ge k$, 
we need:
\begin{equation}
k < \frac{16 c^2 \log T}{9 \delta^2}.
\end{equation}

Therefore, the maximum number of selections in $[t_0, T]$ is bounded by:
\begin{equation}
n_T(s, a) - n_{t_0-1}(s, a) 
\le \left\lfloor \frac{16 c^2 \log T}{9 \delta^2} \right\rfloor + 1
\le \frac{16 c^2 \log T}{9 \delta^2} + 1.
\label{eq:after_t0}
\end{equation}

\textit{Combining the bounds:} Substituting~\eqref{eq:before_t0} 
and~\eqref{eq:after_t0} into~\eqref{eq:count_split}:
\begin{align}
n_T(s, a) 
&\le (t_0 - 1) + \frac{16 c^2 \log T}{9 \delta^2} + 1 \nonumber \\
&= t_0 + \frac{16 c^2 \log T}{9 \delta^2}.
\label{eq:count_intermediate}
\end{align}

\textbf{Step 6: Simplifying the Constant.}

We simplify the bound~\eqref{eq:count_intermediate} by relaxing the constant 
$\frac{16}{9}$ to $16$.

Since $\frac{16}{9} \approx 1.78 < 2 < 16$, we have:
\begin{equation}
\frac{16 c^2 \log T}{9 \delta^2} < \frac{16 c^2 \log T}{\delta^2}.
\end{equation}

Therefore:
\begin{equation}
n_T(s, a) \le t_0 + \frac{16 c^2}{\delta^2} \log T,
\end{equation}
which is exactly the bound~\eqref{eq:selection_app}. This completes the proof.
\end{proof}
\section{Parameter Matching Methodology}
\label{appendix:parameter_matching}
To ensure a fair comparison between tensor-based methods (TEQL, TLR), CUR-based methods (LoRa-VI), and deep reinforcement learning baselines (DQN, SAC), we carefully match the number of learnable parameters across all algorithms.
\subsection{Parameter Formulas}
TEQL and TLR represent the Q-function using a rank-$R$ CP decomposition of a tensor $\mathcal{Q} \in \mathbb{R}^{d_1 \times d_2 \times \cdots \times d_n}$, where each $d_i$ corresponds to the number of discretization buckets. The total parameter count is
\begin{equation}
    P_{\text{CP}} = R \cdot \sum_{i=1}^{n} d_i.
\end{equation}
DQN uses a multilayer perceptron with input dimension $d_{\text{in}}$, hidden layer width $h$, and output dimension $d_{\text{out}}$:
\begin{equation}
    P_{\text{DQN}} = d_{\text{in}} \cdot h + h + h \cdot d_{\text{out}} + d_{\text{out}}.
\end{equation}
SAC employs three such networks plus a temperature parameter: $P_{\text{SAC}} = 3 \times P_{\text{network}} + 1$.
LoRa-VI follows the CUR decomposition approach of~\cite{stojanovic2024lora}, which stores a skeleton of the Q-matrix with $K$ anchor rows and columns:
\begin{equation}
    P_{\text{CUR}} = K \cdot (|\mathcal{S}| + |\mathcal{A}| - K),
\end{equation}
where $|\mathcal{S}| = \prod_i d_i^{\text{state}}$ is the total number of discretized states. To match our parameter budgets, CUR requires substantially coarser discretization than CP: for CartPole, only $3$ buckets per state dimension (versus $10$-$20$ for CP), and for Pendulum, $7$ buckets per dimension (versus $20$ for CP).
\subsection{Parameter Configurations}
Table~\ref{tab:param_all} summarizes the matched parameter configurations across all environments and methods.
\begin{table}[h]
\centering
\small
\caption{Parameter configurations for all environments and methods.}
\label{tab:param_all}
\begin{tabular}{llccc}
\toprule
Environment & Algorithm & Architecture & Params & Ratio \\
\midrule
\multirow{4}{*}{CartPole} 
  & TEQL/TLR & $R=10$, dims=[10,10,20,20,10] & 700 & 1.00$\times$ \\
  & DQN & hidden=[46], in=4, out=10 & 696 & 0.99$\times$ \\
  & SAC & hidden=[38]$\times$3, in=4, out=10 & 703 & 1.00$\times$ \\
  & LoRa-VI & $K=9$, state=[3,3,3,3], $|\mathcal{A}|$=10 & 738 & 1.05$\times$ \\
\midrule
\multirow{4}{*}{Pendulum} 
  & TEQL/TLR & $R=10$, dims=[20,20,10] & 500 & 1.00$\times$ \\
  & DQN & hidden=[38], in=2, out=10 & 504 & 1.01$\times$ \\
  & SAC & hidden=[31]$\times$3, in=2, out=10 & 502 & 1.00$\times$ \\
  & LoRa-VI & $K=10$, state=[7,7], $|\mathcal{A}|$=10 & 490 & 0.98$\times$ \\
\midrule
\multirow{4}{*}{Highway} 
  & TEQL/TLR & $R=20$, dims=[20$\times$9, 5] & 3,700 & 1.00$\times$ \\
  & DQN & hidden=[246], in=9, out=5 & 3,695 & 1.00$\times$ \\
  & SAC & hidden=[82]$\times$3, in=9, out=5 & 3,706 & 1.00$\times$ \\
  & LoRa-VI & \multicolumn{3}{c}{infeasible (see below)} \\
\bottomrule
\end{tabular}
\end{table}
For Highway, CUR-based LoRa-VI becomes infeasible. The 9-dimensional state space yields $|\mathcal{S}| = \prod_i d_i$, which grows exponentially with dimensionality. Even with the coarsest discretization of 2 buckets per dimension, the state space contains $|\mathcal{S}| = 2^9 = 512$ states, resulting in $P_{\text{CUR}} = 20 \times (512 + 5 - 20) = 9{,}940$ parameters, far exceeding the budget of $3{,}700$. This fundamental limitation demonstrates the advantage of CP decomposition, which exploits the factored structure of the state space to achieve parameter complexity of $O(\sum_i d_i)$ rather than $O(\prod_i d_i)$.
\textbf{Remarks.} Hyperparameters such as batch size, buffer size, and learning rate do not contribute to the parameter count. Target networks in DQN and SAC are copies of the main networks and thus counted only once. By equalizing parameters across methods, observed performance differences reflect algorithmic properties rather than model capacity.
\end{document}